\documentclass[10pt,twocolumn,letterpaper]{article}

\usepackage{iccv}              

\definecolor{iccvblue}{rgb}{0.21,0.49,0.74}
\usepackage[pagebackref,breaklinks,colorlinks,allcolors=iccvblue]{hyperref}

\usepackage{mdframed}

\newmdenv[
  backgroundcolor=gray!20,   
  linecolor=gray!20,         
  innerleftmargin=10pt,      
  innerrightmargin=10pt,
  innertopmargin=10pt,
  innerbottommargin=10pt
]{mytextbox}

\newcommand{\ours}{Accordion}

\def\paperID{7413} 
\def\confName{ICCV}
\def\confYear{2025}

\title{Rethinking Layered Graphic Design Generation with a Top-Down Approach}

\author{Jingye Chen$^{1,2}$, Zhaowen Wang$^{2}$, Nanxuan Zhao$^{2}$, Li Zhang$^{2}$, Difan Liu$^{2}$, Jimei Yang$^{3}$, Qifeng Chen$^{1}$ \\
$^{1}$HKUST \;\;\;\;\;\;\; $^{2}$Adobe Research  \;\;\;\;\;\;\; $^{3}$Runway\\
\text{qwerty.chen@connect.ust.hk \;\;  zhawang@adobe.com \;\; cqf@ust.hk}
}

\begin{document}
\maketitle

\begin{abstract}
Graphic design is crucial for conveying ideas and messages. Designers usually organize their work into objects, backgrounds, and vectorized text layers to simplify editing. However, this workflow demands considerable expertise. With the rise of GenAI methods, an endless supply of high-quality graphic designs in pixel format has become more accessible, though these designs often lack editability. Despite this, non-layered designs still inspire human designers, influencing their choices in layouts and text styles, ultimately guiding the creation of layered designs.
Motivated by this observation, we propose \textbf{\ours{}}, a graphic design generation framework taking the \textbf{first attempt} to convert AI-generated designs into editable layered designs, meanwhile refining nonsensical AI-generated text with meaningful alternatives  guided by user prompts. It is built around a vision language model (VLM) playing distinct roles in three curated stages: (1) reference creation, (2) design planning, and (3) layer generation. For each stage, we design prompts to guide the VLM in executing different tasks.
Distinct from existing bottom-up methods (\textit{e.g., COLE and Open-COLE}) that gradually generate elements to create layered designs, our approach works in a top-down manner by using the visually harmonious reference image as global guidance to decompose each layer. Additionally, it leverages multiple vision experts such as SAM and element removal models to facilitate the creation of graphic layers. 
We train our method using the in-house graphic design dataset Design39K, augmented with AI-generated design images coupled with refined ground truth created by a customized inpainting model.
Experimental results and user studies by designers show that \ours{} generates favorable results on the DesignIntention benchmark, including tasks such as text-to-template, adding text to background, and text de-rendering, and also excels in creating design variations.
\end{abstract}

\begin{figure}[t]
\centering
\includegraphics[width=0.475\textwidth]{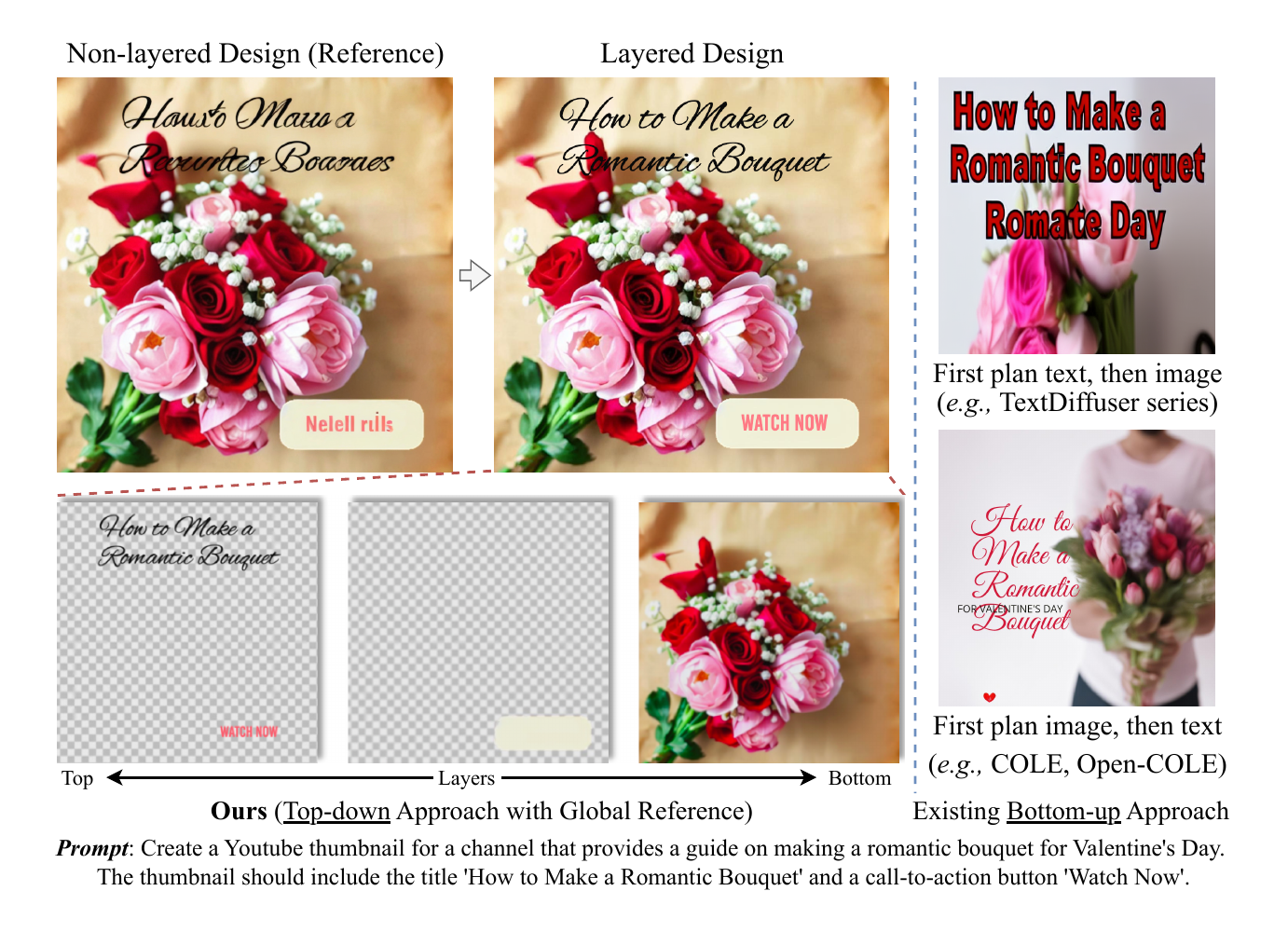}
\caption{\textbf{Left:} We convert AI-generated non-layered design reference images to layered designs by extracting background, objects, and vectorized text layers in a top-down manner with optional further refinement. The layered representation facilitates easier editing. We include a gallery for more visualizations in Appendix A. \textbf{Right:} Existing bottom-up design approaches progressively generate elements on images in a predefined order, yet they lack a visual reference as the overall plan from the start to harmoniously orchestrate each element, often resulting in visual conflicts. For instance, text may occupy too much space, leaving inadequate room for the background, or causing overlaps between text regions.}
\label{fig:teaser}
\end{figure}

\section{Introduction}
\label{sec:intro}

Graphic design is an important media format for modern visual communication. Specifically, graphic design is intrinsically constructed in a layered format, including distinct layers for foreground objects, background, and vectorized text. This structured layering allows for editability and customization. Despite the significant utility, creating layered graphic designs is a prohibitive task for most people due to the need for design expertise and huge effort. With the aid of image GenAI models, more design images have become available in rasterized pixel format. While they are visually compelling, they inherently lack editability. Even for simple operations such as horizontal flipping, text becomes unreadable since they are not separated from the background or other elements. Although users can employ some image editing tools \cite{chen2024textdiffuser,jia2024designedit,hertz2022prompt,he2024llms} to modify the attributes of elements, such an approach is inconvenient compared to operations directly applied on the layer representation.

Nevertheless, we still believe rasterized designs are of great value for creating layered designs, by realizing that human designers naturally use rasterized designs from textbooks or other sources as references to get inspiration in their layered design workflows. For instance, designers will use them in the beginning to explore suitable layouts, deciding where to place objects and what style of typography to use to achieve visual harmony. Given this practice, we aim to leverage rasterized designs as global references to create editable multi-layer designs, as illustrated in Figure \ref{fig:teaser} (left).

\begin{figure}[t]
\centering
\includegraphics[width=0.5\textwidth]{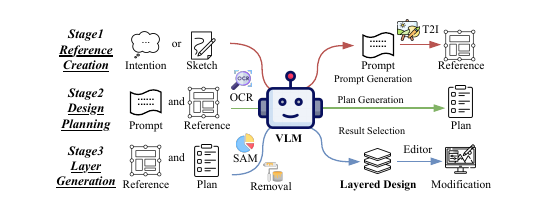}
\caption{Overview of the proposed \ours{} built around a VLM. It consists of three stages for constructing the design reference, plan, and layers respectively. Multiple vision experts (\textit{e.g.,} OCR and SAM) are employed to create layered designs.}
\label{fig:introduction}
\end{figure}

Based on this motivation, we introduce \textbf{\ours{}}\footnote{Our method is named ``Accordion'' because it unfolds rasterized designs into layered designs, similar to how an accordion expands. Besides, the proposed Accordion framework harmoniously integrates each element within a layered design, much like orchestrating every note harmoniously in a musical score at the concert.}, a framework as demonstrated in Figure \ref{fig:introduction} built around a vision language model VLM consisting of three stages: (1) \textit{reference creation}, (2) \textit{design planning}, and (3) \textit{layer generation}, while the VLM plays different roles in these stages. Notably, the framework can leverage a diverse and unlimited range of AI-generated references. It also offers the flexibility to start from the second stage when users explicitly provide designs as references. Besides, the VLM uses some vision experts in the design process. For example, SAM \cite{kirillov2023segment} and removal models \cite{rombach2022high} are used for element extraction and background filling. To the best of our knowledge, we take the \textbf{first attempt} to convert AI-generated designs to editable layered designs.

As shown in Figure \ref{fig:teaser} (right), in contrast to earlier bottom-up works, including text rendering methods such as TextDiffuser \cite{chen2024textdiffuser}\footnote{Text rendering methods usually require an additional text segmentation model and inpainting model for image layering.} that specifies the location of text before designing the background, and unlike COLE \cite{jia2023cole} that starts with the background to design text in the visual domain, our method starts with a global reference image representing the target design as a whole. This initial reference image globally orchestrates the layout and visual properties of various elements, ensuring overall visual harmony and preventing the conflicts that may arise in the step-by-step generation where earlier design decisions may not work well with later ones, as exemplified by generating text without leaving too much space for the foreground objects.

\ours{} is trained using the in-house graphic design dataset Design39K. We innovatively use an inpainting model to mimic AI-generated design images and also obtain the refined text ground truth to construct the pairs.
Experimental results and visualizations show the strong design capabilities of our method, especially its superior performance across various tasks on the DesignIntention benchmark \cite{jia2023cole}, including text-to-template, adding text to background, and text de-rendering. Additionally, we explore the potential of Accordion to facilitate creative design variation, including using upstream generative models to modify references, applying inference time variations, and utilizing downstream generative models for further variations based on the extracted layers. Overall, our contributions can be summarized as follows:

\begin{itemize}
    \item We rethink the layered design process by adopting a top-down approach, using the global reference to decompose each layer ensuring that all elements are harmonious. We take the first attempt to convert AI-generated designs into editable and practical layered designs.
    \item Accordion achieves superior performance on the DesignIntention benchmark in tasks such as text-to-template, adding text to backgrounds, and text de-rendering.
    \item Accordion enables a variety of variations and benefits from integration with existing generative models.
\end{itemize}

\section{Related Work}
\paragraph{Layered Design Generation}
There are a few investigations that focus on generating layered designs \cite{jia2023cole,inoue2024opencole,shimoda2021rendering,pu2025art,chen2025posta} considering the need for editability and customization. For instance, COLE \cite{jia2023cole} starts from a brief user-provided prompt, employing multiple large language models (LLMs) and diffusion models to generate each element within the design. Even though COLE uses language to comprehensively plan the design, it still visually constructs the design step-by-step, starting from the background, then generating objects, and finally the text. This sequential approach may lead to visual conflicts such as failing to allocate sufficient or suitable space for text or objects when generating background, often resulting from the lack of a global visual impression in mind. Open-COLE \cite{inoue2024opencole} adheres to the architecture of COLE but incorporates certain simplifications, such as omitting the object generation stage. Additionally, commercial tools such as CanvaGPT \cite{canvagpt} utilize preset templates to adapt user prompts to new layered
designs as noted in \cite{jia2023cole}. However, the diversity of designs is limited by the size of template libraries. We notice that a few methods \cite{zhang2024transparent,zhang2023text2layer,tudosiu2024mulan} are designed for the layer generation of natural images. They are unsuitable for design images due to differences such as the absence of text layers in natural images. Another line of research involves predicting the attributes of partial elements given any existing elements, which are then combined together to create a layered design. Some works focus on predicting the layout of elements within a design, typically emphasizing the importance of proper alignment and the avoidance of overlaps to enhance visual aesthetics \cite{shabani2024visual,jyothi2019layoutvae,li2019layoutgan,gupta2021layouttransformer,hsu2023posterlayout,seol2024posterllama,li2023planning,li2021harmonious,zhu2024automatic,zheng2019content,zhou2022composition,chai2023two,horita2024retrieval,jiang2022coarse,cheng2024graphic,liang2024textcengen,zhang2023layoutdiffusion,biswas2021docsynth,yamaguchi2021canvasvae,yu2022layoutdetr,kikuchi2024multimodal,inoue2023towards,zhao2018modeling,lin2023autoposter,biswas2024docsynthv2}.

Overall, we first explore how to construct layered designs powered by the generative model and VLM, distinguishing it from existing methods.

\paragraph{Non-layered Design Generation}

Existing GenAI methods are capable of creating an unlimited number of non-layered design images. Methods like Stable Diffusion \cite{rombach2022high,podell2023sdxl,esser2024scaling} has demonstrated this capability of design image generation guided by user prompts. However, these methods often suffer from text rendering errors \cite{daras2022discovering}, which can significantly impact the usability and aesthetic of the generated images. To alleviate the text rendering issue, some works incorporate control over text areas during the image generation process \cite{chen2024textdiffuser,chen2023textdiffuser,tuo2023anytext,ma2024glyphdraw2,ma2023glyphdraw,zhang2024brush,zhao2023udifftext,ji2023improving,liu2024glyph,liu2024glyph2,gao2023textpainter,zhangli2024layout,zhao2024harmonizing,zeng2024textctrl,chen2025postercraft,lan2025flux}. While these methods improve the accuracy of text rendering, the final images are rasterized in pixel format, which prevents users from making edits easily.

Overall, despite the limitations of non-layered designs, we suppose these images are visually compelling and offer valuable references for the creation of layered designs.

\section{Methodology}
We sequentially detail the three stages of the proposed \ours{} framework as illustrated in Figure \ref{fig:introduction}, explaining the different roles the VLM plays in each stage. Note that we use AI-generated images as references for illustrative purposes. \ours{} can also start from the later two stages if users explicitly provide other types of references such as an existing design or a background layer. 

\paragraph{Problem formulation.} Starting from an initial short intention $I$ or sketch image $S$, we obtain an intermediate reference image $R$ as global guidance, and expect output as a set of layers $\{O_{1}, O_{2}, \ldots, O_{N}, B, T\}$, where $O_{*}$ represents object layers, where $N$ is the total amount of foreground object layers. $B$ is a background layer, and $T$ is a vectorized text layer. Finally, these layers are stacked in the predicted order to form a layered design.

\begin{figure*}[t]
\centering
\includegraphics[width=1.0\textwidth]{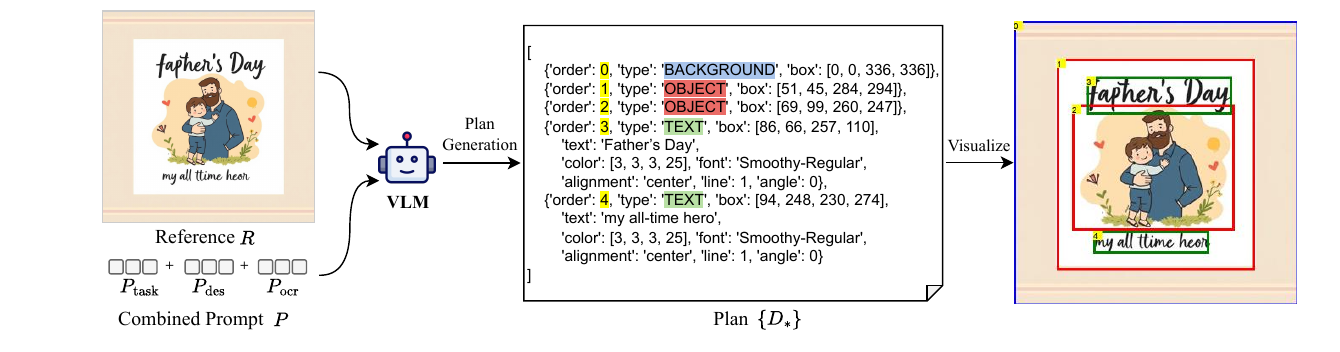}
\caption{In Stage2 design planning, VLM processes the reference image alongside a combined prompt to generate a comprehensive design plan. This plan includes detailed information about each design element, including the background, objects, and text.}
\label{fig:stage2}
\end{figure*}

\begin{figure*}[t]
\centering
\includegraphics[width=1.0\textwidth]{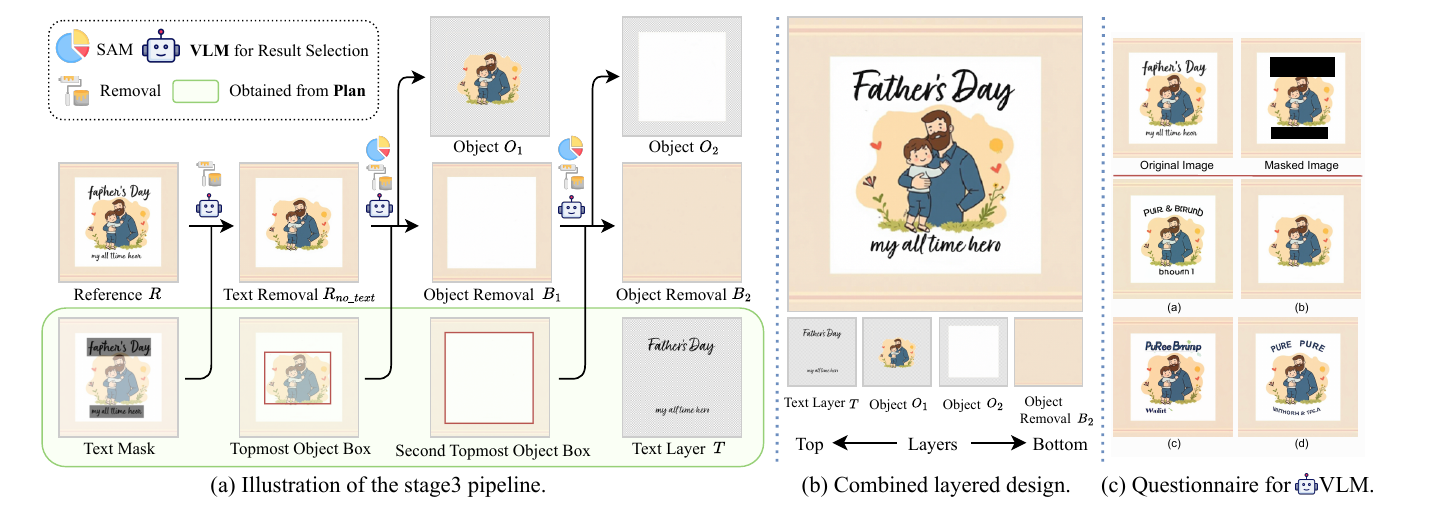}
\caption{Overview of the Stage3 layer generation. (a) Guided by the plan, the reference image is processed first with text removal, and then with progressive foreground object extraction using SAM and an object removal model, obtaining the background image in the end. (b) The text, objects, and background are stacked into a layered design. (c) The VLM conducts result selection using questionnaires throughout this process for better removal results.}
\label{fig:stage3}
\end{figure*}

\subsection{Stage1: Reference Creation}
In this stage, our objective is to generate an image that serves as a global reference for the later two stages.
\ours{} is designed to accommodate a variety of inputs, including user-provided short intentions $I$ or sketch drafts $S$, thus catering to diverse user preferences and needs.
To facilitate this, we employ the VLM for \textbf{prompt generation}. As mentioned in COLE \cite{jia2023cole}, users may provide only short intentions for simplicity, such as ``\textit{create a poster for Father's day}''. We use in-context learning by supplying the VLM with multiple examples for prompt enhancement following Open-COLE \cite{inoue2024opencole}. This approach enables the VLM to generate detailed descriptive prompts $P_{des}$ such as \textit{``A father embraces his child in the center, surrounded by the text Father's Day and My All Time Hero.''}. It is observed that detailed and lengthy prompts can facilitate the creation of images with rich detail \cite{chen2023pixart} and significantly improve text quality \cite{chen2023textdiffuser} for better reference. If users wish to provide more layout constraints, they can use sketch drafts. In such cases, the VLM is instructed to give detailed descriptions of the depicted objects, accurately depicting object positions and filling text in suitable areas. We show the prompt templates and visualizations in Appendix B. Subsequently, these generated prompts are fed into a text-to-image (T2I) model to generate images as references $R$ to be used in the next stage.

\subsection{Stage2: Design Planning}

In this stage, we aim to derive a design plan based on the rasterized reference image. This plan should detail the placement order of objects within the image to prepare for subsequent extraction, and provide the rendering attributes of text to facilitate the construction of vectorized text layers.

We employ VLM for \textbf{plan generation}. As illustrated in Figure \ref{fig:stage2}, the VLM processes the reference image alongside a combined prompt $P$, which is concatenated by a predefined task description $P_{\text{task}}$, a description $P_{\text{des}}$, and OCR string $P_{\text{ocr}}$.
The description is derived from the first stage detailed prompt, which is crucial for adapting textual content within the design plan, particularly for refining any nonsensical text produced by generative models.
For OCR string, we use the text detection results of GenAI images, as the nonsensical content offers limited reference value. We show some cases of combined prompts in Appendix C.

As for the output, the VLM generates a sequence of dictionaries $\{D_{*}\}$, each representing the attributes of elements in the image, arranged in a bottom-to-top order to facilitate further object extraction. The output includes bounding boxes for both background and foreground objects and detailed text attributes such as bounding boxes, content, color, font, alignment, line count, and angle. Box coordinates are normalized to the range [0, 336]. For the color attributes (R, G, B, A), we map the [0, 255] range to [0, 25] to facilitate learning inspired by COLE \cite{jia2023cole}.  Extending these attributes is straightforward and can be accomplished by incorporating additional fields into the training dataset. So far, this design plan forms the foundation for the next stage.

\begin{figure*}[t]
\centering
\includegraphics[width=1.0\textwidth]{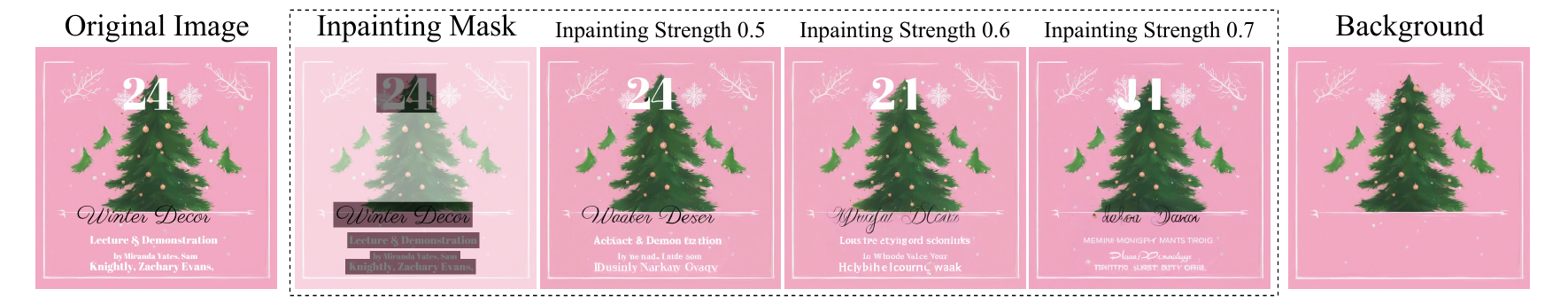}
\caption{For the same design, we utilize three types of references for the VLM, corresponding to three distinct training tasks: original designs, designs featuring nonsensical text, and designs with the text removed. We employ an inpainting model of varying strengths to generate images containing nonsensical text, where higher inpainting strength results in greater distortion of text details.}
\label{fig:dataset}
\end{figure*}

\subsection{Stage3: Layer Generation}
In this stage, our goal is to construct the layered design. As illustrated in Figure \ref{fig:stage3}, our approach is to extract and remove elements from the reference image, and then stack them back together to construct the final layered design. Although De-Render \cite{shimoda2021rendering} also possesses the capability to extract layers from design images, it cannot extract object layers and lack the capability to refine nonsensical words. 

Our initial step is to use a text removal model to erase text from the reference image $R$, recognizing that text regions are commonly placed on the top layer. Note that in the generated plan $\{D_{*}\}$, each text's bounding box is already included, which we can use as a condition to remove the text. Besides, the vectorized text layer $T$ can be obtained from the plan, detailing the attributes of the text. Based on the text removal result $R_{no\_text}$, our next focus moves to object removal, where we sequentially extract the topmost element according to the order outlined in the design plan. The object removal can be executed iteratively if multiple objects are detected in the design plan. In detail,
for the $n$th object extraction, we use the SAM on the image $B_{n-1}$ (when $n=1$, the image $R_{no\_text}$ is used instead), which is conditioned on the bounding box from the design plan, to extract the foreground object $O_{n}$ and its mask. 
The mask and the intermediate image are fed into an object removal model to remove the foreground object and obtain the intermediate background $B_{n}$. The last predicted background is used as the final background $B$.

It is noticed that the removal model sometimes generates diverse results, not all of which are satisfactory. To ensure consistent quality, we design a questionnaire that enables the VLM to conduct \textbf{result selection}, as illustrated in Figure \ref{fig:stage3} (c). The top row shows the original image alongside the masked image, where the text is masked with text boxes for the text removal task, and the object is masked using the SAM segmentation map for the object removal task. During the training phase, we present the VLM with the ground truth of the removal alongside three generated removal results, training it to select the highest quality option. In the inference stage, we generate four removal results for the VLM to choose the best one. We showcase the task prompt and some samples in Appendix D. So far, all the extracted objects $\{O_{1}, O_{2}, \ldots, O_{n}\}$ and the background $B$ are combined with the text layer $T$ rendered from the design plan to create a layered design.

\section{Experiments}

\subsection{Implementation Details}

\begin{figure*}[t]
\centering
\includegraphics[width=1.0\textwidth]{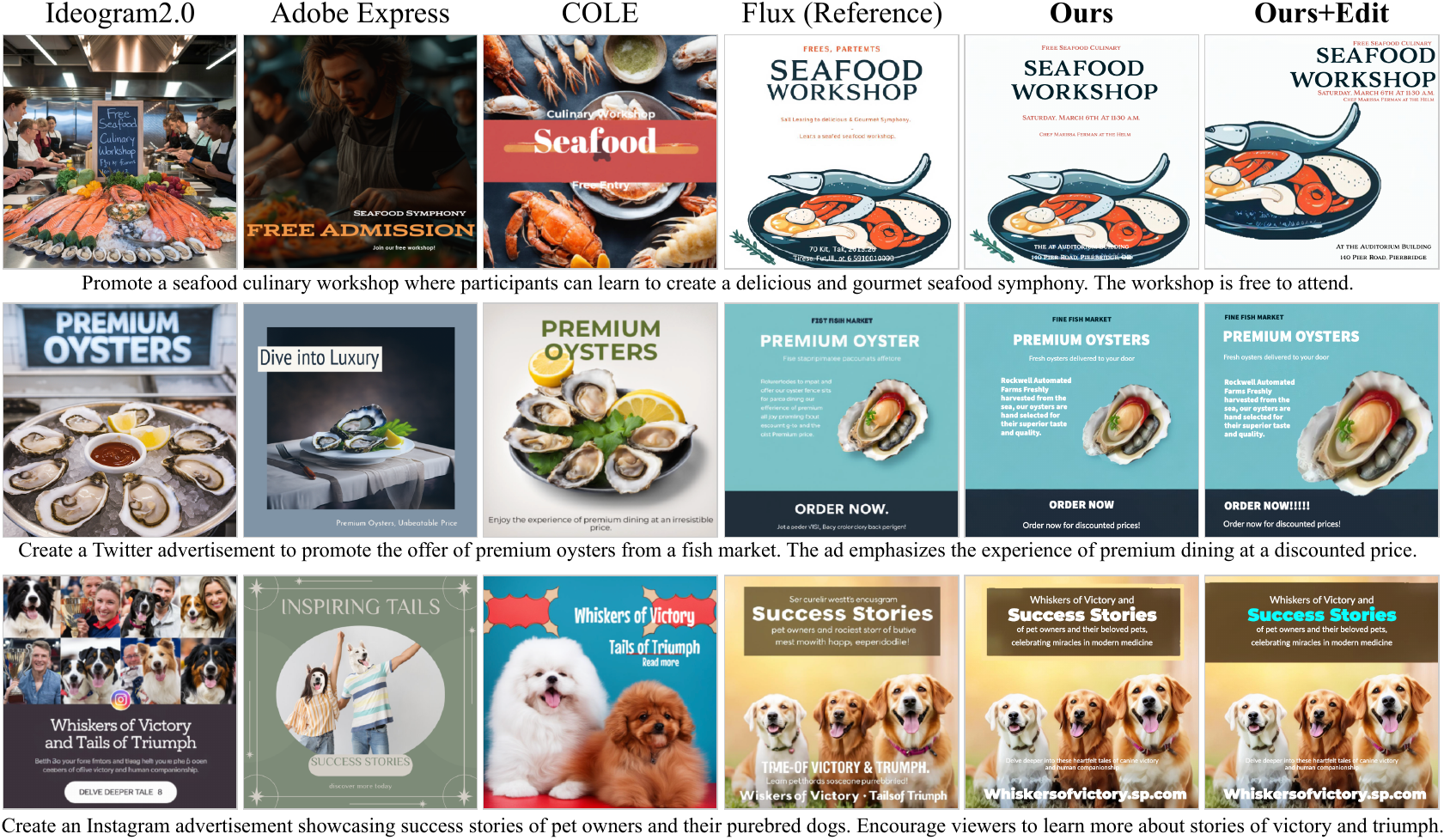}
\caption{Qualitative results of text-to-template show that our method uses references generated by Flux to create layered designs where elements appear more harmonious, and supports layered editing compared with Ideogram2.0. Intentions are shown below each image.}
\label{fig:text-to-template}
\end{figure*}

\textbf{Dataset.} We employ an in-house layered design dataset \textbf{Design39K}, with 39,233 samples for training and 492 samples for validation. This dataset comprises a diverse array of designs, including posters, book covers, advertisements, etc. Each sample is accompanied by a description. It is easy to extract layer information from these designs, including text, objects, and backgrounds, along with various attributes. We present more details about the dataset in Appendix E.
As shown in Figure \ref{fig:dataset}, to enrich the reference sources, we incorporate the following types of designs for training: (1) \textit{Original designs}. We wish the model to conduct text de-rendering by directly parsing these original designs; (2) \textit{Designs with nonsensical text}. We employ the Stable Diffusion 1.5 inpainting model \cite{rombach2022high} to inpaint text areas with inpainting strength randomly set between 0.5 and 0.7, leading to the generation of nonsensical text by the model. This range is selected because strength outside this range can lead to either insufficient or excessive inpainting changes, which hinder effective training. Please note that in some cases inpainting results may not strictly maintain the original text style, with potential variations in color and font. We consider this acceptable as it enables the model to use references while generating creative variations. The goal is to enable our model to effectively handle GenAI designs; (3) \textit{Background without text}. By removing all text from a design and using only the background image as a reference, we challenge the model to add text in appropriate context and locations. We organize the elements of the design into a list of dictionaries and convert them into a string format for training the VLM. We employ the same training objectives for the three aforementioned reference types. Besides, to train the VLM for result selection, for both text and object removal tasks, we use the removal model to generate three different results. These results are then combined with the ground truth and randomly shuffled to construct the questionnaire dataset. In total, we have prepared 156,932 samples, consisting of 39,233 training samples for each of the three different types of references and the questionnaire. 

\textbf{Selection of vision expert models.}
For the Text-to-Image (T2I) model, we utilize Flux for its proficiency in generating high-quality references. Additionally, we use PaddleOCR as the OCR tool. SAM \cite{kirillov2023segment} and inpainting ControlNet \cite{zhang2023adding} for Stable Diffusion 1.5 \cite{rombach2022high} with the prompt ``nothing in the image'' are used in the layer generation stage. Importantly, these models are modular and can be replaced when more advanced alternatives are available. We detail each expert model in Appendix F.

\textbf{Training, inference, evaluation, and visualization.} We use the vision language model LLaVA-1.5-7B \cite{liu2024visual} as the cornerstone of the proposed \ours{}. The model is trained on the aforementioned 156,932 samples using LoRA \cite{hu2021lora} with learning rate 2e-4 for 6 epochs, conducted on 8 $\times$ 80G A100 GPUs for 36 hours. The reference image is scaled with the longer side set to 336 pixels following \cite{liu2024visual} for the VLM. The removal models operate at a resolution of 512$\times$512, which is the size of the final output. During inference, the average generation time per sample is 36.7 seconds (7, 19.5, 10.2 seconds on average for the three stages, respectively). For evaluation, we use the DesignIntention benchmark \cite{jia2023cole} which provides 500 detailed prompts across various design domains. We visualize the layered design through a Streamlit HTML frontend, with details provided in Appendix G.

\begin{figure*}[t]
\centering
\includegraphics[width=1.0\textwidth]{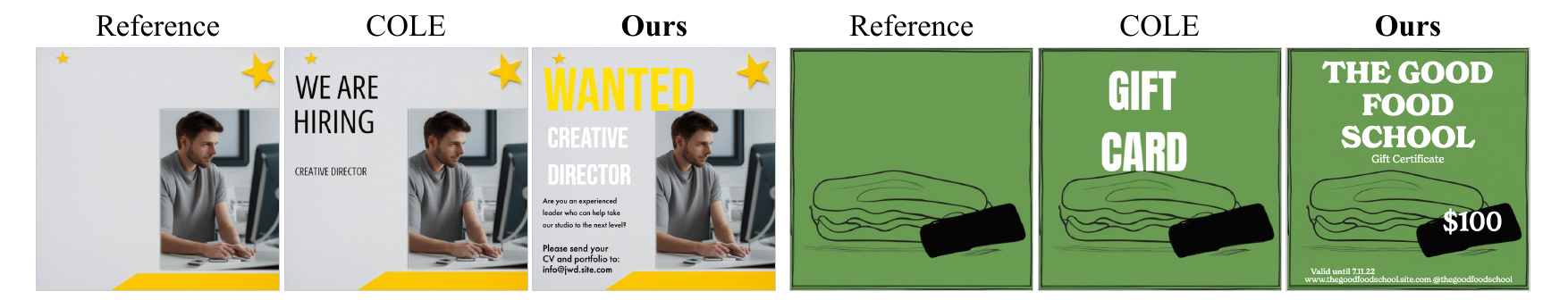}
\caption{Qualitative results of adding text to the background show that our method not only places text harmoniously in terms of style and layout but also effectively utilizes space to create more complex and aesthetically pleasing designs than COLE.}
\label{fig:backgroundadd}
\end{figure*}

\begin{figure*}[t]
\centering
\includegraphics[width=1.0\textwidth]{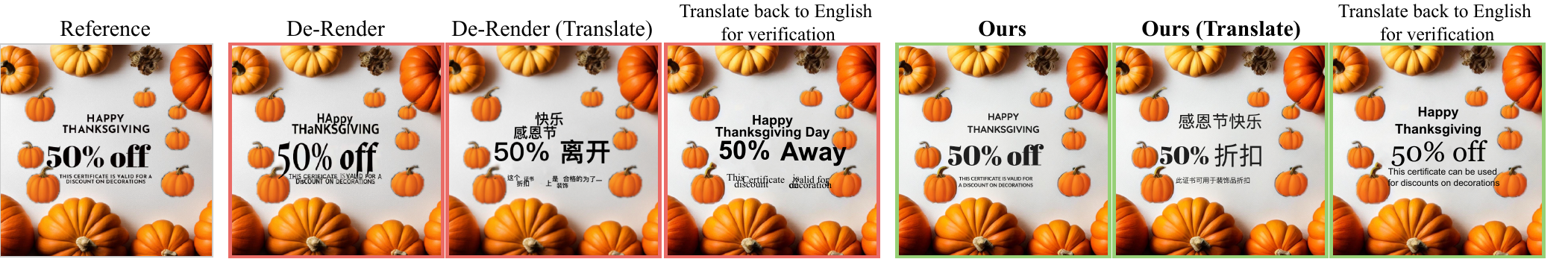}
\caption{Qualitative results of text de-rendering. Our method groups text at the paragraph level to predict their attributes, thereby enhancing visual harmony. Through the translation to Chinese task, we better maintain semantic coherence, style harmony, and spatial alignment compared with De-Render, which predicts the attributes of each individual word, resulting in an overall appearance that lacks harmony. We also translate the design from Chinese back to English to verify its consistency with the original design. Our approach significantly surpasses the original in terms of semantic and layout consistency.}
\label{fig:derender}
\end{figure*}

\subsection{Experimental Results}

\textbf{Text-to-template.} As shown in Figure \ref{fig:text-to-template}, we generate reference images using Flux \cite{flux} and compare them with commercial tools such as Ideogram2.0 and Adobe Express, as well as with images generated by COLE \cite{jia2023cole}. Some samples are shown in Figure \ref{fig:text-to-template}. It is noteworthy that although the images produced by Ideogram2.0 appear fancy, they lack editability, and some text areas exhibit artifacts.
Adobe Express produces vector designs with professional styles, but the generated text does not always match the input query closely. For example, in the second row, we intend to generate an advertisement for an ``oyster discount'', but the output contains a text saying ``dive into luxury''. COLE can generate layered designs, but the elements lack visual harmony. For example, in the first row, the placement of ``Culinary Workshop'' and ``Free Entry'' looks unnatural; and in the third row, the positioning of two red objects conflicts with the foreground text. Our method uses Flux to create references. We observe that due to the exceptional generative capabilities of Flux, the rasterized images it produces feature visually harmonious elements, despite the presence of nonsensical text. Then our method extracts objects from these references, learns the text style, and improves upon nonsensical text areas in the references, demonstrating superior performance. We also showcase editing results in the last column, including operations on text and objects. It is worthy of note that our method is adaptable to reference images generated by various models, such as substituting Flux with SD3 \cite{esser2024scaling} or future advanced models. In contrast, COLE relies on fine-tuning a base generative model to achieve layered designs, necessitating retraining to adapt to newer generative models, which diminishes its compatibility. Furthermore, the proposed \ours{} supports the generation of designs with varying aspect ratios, whereas COLE is limited to producing only square designs. These results are detailed in the Appendix H.

We validate the quantitative results using the protocol provided by COLE, which evaluates performance across five metrics. These metrics are scored via preset prompt queries to GPT-4V, with scores ranging from 0 to 10 and higher scores indicate better performance. Our comparisons are primarily with academic methods following \cite{inoue2024opencole} because commercial tools typically lack open-source code or accessible APIs for large-scale testing. As shown in Table \ref{tab:model_comparison}, our method achieves the highest average score (6.5) compared to existing layered approaches such as COLE and Open-COLE. It is worth noting that our method outperforms COLE, despite the fact that COLE uses a 100K in-house dataset for training, which is far more extensive than our dataset of 39K. In addition, we specifically evaluate the content relevance of images generated by Flux and our method. The experimental results show that our method outperforms Flux by 0.4 in this metric (7.0 for Flux \textit{v.s.} 7.4 for ours). This improvement demonstrates that our method refines the image by replacing nonsensical text with relevant information, thereby enhancing the content relevance of text layers to backgrounds. We also employ an aesthetic score for evaluation. Specifically, we utilize LAION's aesthetic tool to assess the aesthetic quality of the designs. Our method achieved a high score of 4.98, outperforming COLE's method, which scored 4.72.

\textbf{Adding text to background.} 
Since COLE provides the SVG-format results for the DesignIntention benchmark \cite{jia2023cole}, we can easily remove the text to obtain the background. As illustrated in Figure \ref{fig:backgroundadd}, we compare our method with the results from COLE. The results demonstrate that our method achieves good harmony in both layout and text style. In addition, we observe that COLE tends to produce layouts that are relatively simple, often avoiding longer sentences. Through analysis, it is evident that our model is able to generate text with more informative content. We conclude that the reason for this issue is that COLE pre-determines the text content without considering the available space in the background image, potentially resulting in a disruption of visual harmony. Our samples show a greater overall text length, averaging 61.7 characters compared to 42.3 characters for COLE (approximately 1.5 times longer). In other words, our approach can fully utilize empty space in the background to increase information density and enhance overall aesthetics. This makes our model especially well-suited for scenarios like detailed reports and comprehensive advertisements. We quantitatively compare our results with COLE using GPT-4V, where images generated by both methods are concatenated horizontally and shown to GPT-4V to assess which one has higher quality. The task prompt is provided in Appendix I. The results show a preference for our model on 52\% of all the samples compared to 48\% for COLE.

\begin{table}[t]
\caption{Quantitative results for the comparison of layered design methods using the DesignIntention benchmark. The five metrics are: (i) design and layout, (ii) content relevance, (iii) typography and color, (iv) graphics and images, and (v) innovation. $^{*}$Open-COLE cannot generate the foreground object layers. The proposed \ours{} achieves the best average performance.}
\vspace{-0.6cm}
\label{tab:model_comparison}
\begin{center}
\resizebox{0.48\textwidth}{!}{
\begin{tabular}{lcccccc}
\toprule
Methods & (i) & (ii) & (iii) & (iv) & (v) & Avg. \\
\midrule
COLE \cite{jia2023cole} & 6.0 & 6.9 & 5.7 & 6.2 & 5.1 & 6.0 \\
\textcolor{gray}{Open-COLE$^{*}$} \cite{inoue2024opencole} & 6.3 & 7.0 & 5.6 & 7.1 & \textbf{5.3} & 6.3 \\
\textbf{Accordion (Ours)} & \textbf{6.7} & \textbf{7.4} & \textbf{6.1} & \textbf{7.3} & 5.1 & \textbf{6.5} 
\\
\bottomrule
\end{tabular}}
\vspace{-0.5cm}
\end{center}
\end{table}

\begin{figure*}[t]
\centering
\includegraphics[width=0.98\textwidth]{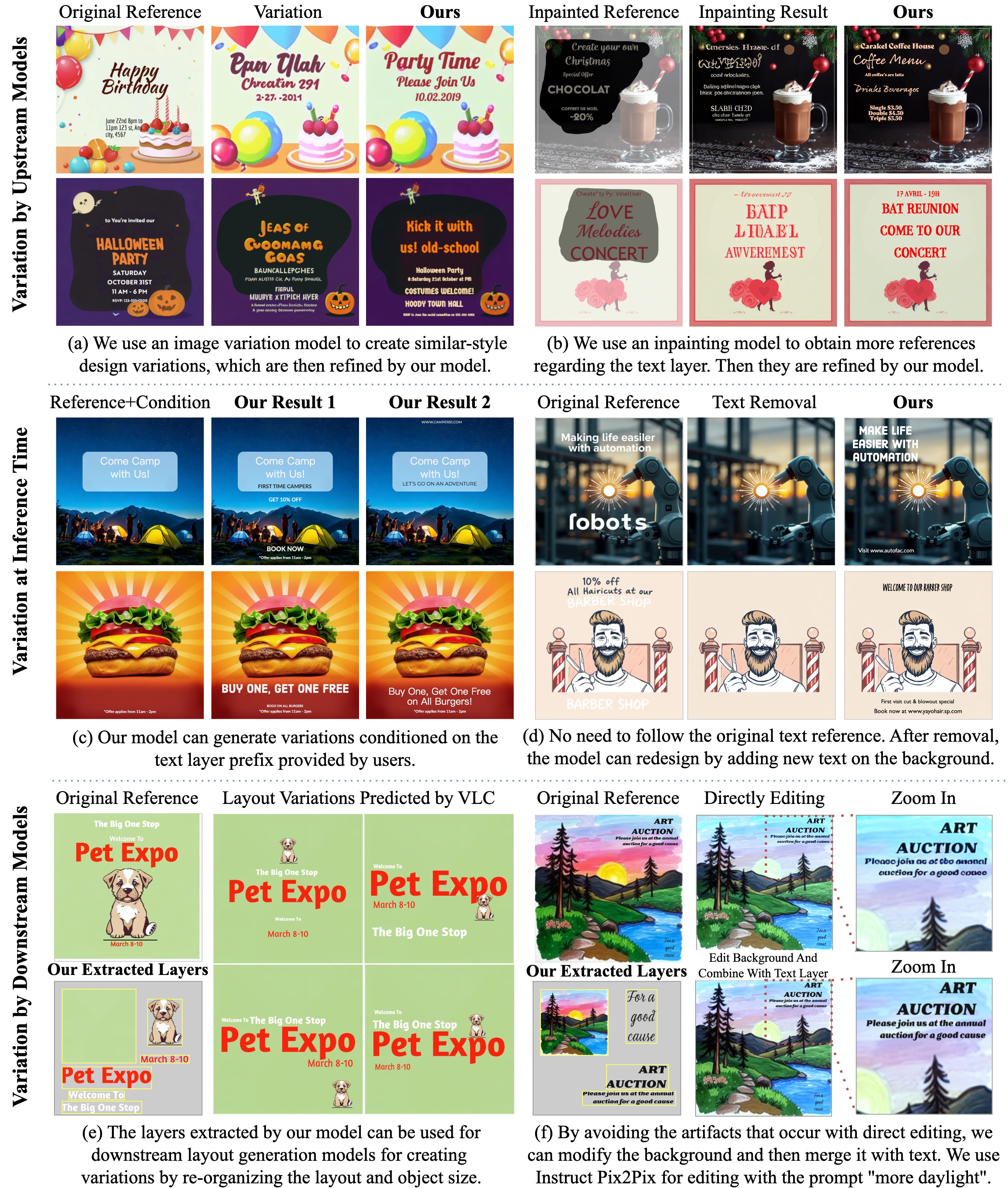}
\caption{Our methods can support the design variations.}
\label{fig:variation}
\end{figure*}


\textbf{De-rendering.} 
In Figure \ref{fig:derender}, we validated the text de-rendering capability of our model in comparison with De-Render \cite{shimoda2021rendering}. Note that our method treats paragraph-level text as a single entity. One advantage of paragraph-level representation is the uniformity of style, a feature that De-Render struggles to achieve. In contrast, De-Render predicts a style for each word individually, which can lead to a visually disorganized appearance. Moreover, we showcase our advantage with a translation application. By grouping words at the paragraph level, our approach effectively considers the coherence of sentence semantics during translation, ensuring consistent style and alignment of adjacent words. Since De-Render operates at the word level, it lacks overall contextual information in translation application, leading to disorganized layouts with overlapping text. 

Another critical evaluation metric is the ability to de-render the correct number of layers from a rasterized design. We test on the detection of layer quantities using the Design39K evaluation set. The mean absolute error (L1 loss) between detected values and ground truth, is 0.494 for text layers and 0.274 for object layers. This indicates that the detection of layer numbers is relatively accurate.

\textbf{Design variation creation.} As shown in Figure \ref{fig:variation}, we demonstrate three methods based on our approach for creating design variations under three categories: (a-b) \textit{Variation by upstream models} based on an image variation model \cite{xu2023versatile}. Typically, image editing models such as variations models or inpainting models tend to degrade the text regions within images. Our method is capable of refining these areas. (c-d) \textit{Variation at inference time}. The VLM can generate different prediction branches based on the existing text on the background, corresponding to various designs. (e-f) \textit{variation by downstream models} based on VLC \cite{shabani2024visual}. Our model is capable of decomposing existing designs for downstream layout generation models. Overall, these methods demonstrate the versatility and critical role of our approach in the design workflow.



\subsection{User Studies by Designers}
To evaluate the \textit{editability} of the layered design and to differentiate it from COLE and Open-COLE, which rely solely on image-level evaluation, we enlisted 29 designers to assess 30 cases. In the text-to-template editability evaluation, 73.5\% of the cases were deemed superior to those of COLE. Furthermore, in the sketch-to-description task, 87.2\% of the cases were considered appropriate. These higher scores demonstrate that users can effectively operate the system with straightforward inputs, thereby significantly reducing complexity. Detailed results of the user studies are presented in Appendix J.

\subsection{Ablation Studies}


\noindent \textbf{Multi-task training for VLM.} Here we investigate whether joint training gains benefits or leads to degradation. The results for each task are presented in Appendix K. We observe that joint training and separate training yield comparable average scores, with joint training slightly outperforming by a margin of 0.82\%. We opt for a more compact model architecture and ultimately choose joint training.


\noindent \textbf{Questionnaire selection effect.} We evaluate the efficacy of using questionnaires for the result selection process. The results indicate that for text removal tasks, the PSNR increases from 18.01 to 21.18, and for object removal tasks, it improves from 22.92 to 24.87. It demonstrates the effectiveness of this approach. 


\section{Conclusions and Discussions}
We introduce \ours{}, which takes the first attempt to convert AI-generated designs to editable designs. It has three stages and uses reference images to create layered designs. We have employed plenty of evaluation metrics to assess the performance of the proposed method, including the GPT evalaution score, aesthetic score, as well as the evaluation by human designers to rate the editability. It shows superior performance on the DesignIntention benchmark and we fully explore the potential of Accordion to create the design variations. For the \underline{\textit{limitations}}, the SAM model attained an IOU score of 68.4\%, occasionally facing challenges with hollow or transparent parts in objects and incomplete extractions. These issues often stem from cumulative errors in upstream VLM detection, including misaligned bounding boxes around the objects. Besides, the current approach assumes that the text layer always precedes the object layer in the visual hierarchy, which simplifies the extraction process. Last, the extracted text layer is limited to predefined 2,000 styles and cannot represent freeform styles or text with special visual effects.
For \underline{\textit{future work}}, we plan to extend the JSON structure in the planning stage to incorporate additional attributes. Then we aim to employ VLM as an agent to integrate additional expert models.

{
    \small
    \bibliographystyle{ieeenat_fullname}
    \bibliography{main}

\begin{thebibliography}{65}
\providecommand{\natexlab}[1]{#1}
\providecommand{\url}[1]{\texttt{#1}}
\expandafter\ifx\csname urlstyle\endcsname\relax
  \providecommand{\doi}[1]{doi: #1}\else
  \providecommand{\doi}{doi: \begingroup \urlstyle{rm}\Url}\fi

\bibitem[Biswas et~al.(2021)Biswas, Riba, Llad{\'o}s, and Pal]{biswas2021docsynth}
Sanket Biswas, Pau Riba, Josep Llad{\'o}s, and Umapada Pal.
\newblock Docsynth: a layout guided approach for controllable document image synthesis.
\newblock In \emph{ICDAR}, 2021.

\bibitem[Biswas et~al.(2024)Biswas, Jain, Morariu, Gu, Mathur, Wigington, Sun, and Llad{\'o}s]{biswas2024docsynthv2}
Sanket Biswas, Rajiv Jain, Vlad~I Morariu, Jiuxiang Gu, Puneet Mathur, Curtis Wigington, Tong Sun, and Josep Llad{\'o}s.
\newblock Docsynthv2: A practical autoregressive modeling for document generation.
\newblock \emph{arXiv preprint arXiv:2406.08354}, 2024.

\bibitem[canva(2024)]{canvagpt}
canva.
\newblock Link: https://www.canva.com/magic-write/, 2024.

\bibitem[Chai et~al.(2023)Chai, Zhuang, Yan, and Zhou]{chai2023two}
Shang Chai, Liansheng Zhuang, Fengying Yan, and Zihan Zhou.
\newblock Two-stage content-aware layout generation for poster designs.
\newblock In \emph{ACMMM}, 2023.

\bibitem[Chen et~al.(2025{\natexlab{a}})Chen, Xu, Li, Ren, Ye, Liu, Chen, Zhu, and Wang]{chen2025posta}
Haoyu Chen, Xiaojie Xu, Wenbo Li, Jingjing Ren, Tian Ye, Songhua Liu, Ying-Cong Chen, Lei Zhu, and Xinchao Wang.
\newblock Posta: A go-to framework for customized artistic poster generation.
\newblock In \emph{CVPR}, 2025{\natexlab{a}}.

\bibitem[Chen et~al.(2023)Chen, Huang, Lv, Cui, Chen, and Wei]{chen2024textdiffuser}
Jingye Chen, Yupan Huang, Tengchao Lv, Lei Cui, Qifeng Chen, and Furu Wei.
\newblock Textdiffuser: Diffusion models as text painters.
\newblock In \emph{NeurIPS}, 2023.

\bibitem[Chen et~al.(2024{\natexlab{a}})Chen, Huang, Lv, Cui, Chen, and Wei]{chen2023textdiffuser}
Jingye Chen, Yupan Huang, Tengchao Lv, Lei Cui, Qifeng Chen, and Furu Wei.
\newblock Textdiffuser-2: Unleashing the power of language models for text rendering.
\newblock In \emph{ECCV}, 2024{\natexlab{a}}.

\bibitem[Chen et~al.(2024{\natexlab{b}})Chen, Yu, Ge, Yao, Xie, Wu, Wang, Kwok, Luo, Lu, et~al.]{chen2023pixart}
Junsong Chen, Jincheng Yu, Chongjian Ge, Lewei Yao, Enze Xie, Yue Wu, Zhongdao Wang, James Kwok, Ping Luo, Huchuan Lu, et~al.
\newblock Pixart-alpha: Fast training of diffusion transformer for photorealistic text-to-image synthesis.
\newblock In \emph{ICLR}, 2024{\natexlab{b}}.

\bibitem[Chen et~al.(2025{\natexlab{b}})Chen, Lai, Gao, Ye, Chen, Shi, Shao, Lin, Fei, Xing, et~al.]{chen2025postercraft}
SiXiang Chen, Jianyu Lai, Jialin Gao, Tian Ye, Haoyu Chen, Hengyu Shi, Shitong Shao, Yunlong Lin, Song Fei, Zhaohu Xing, et~al.
\newblock Postercraft: Rethinking high-quality aesthetic poster generation in a unified framework.
\newblock \emph{arXiv preprint arXiv:2506.10741}, 2025{\natexlab{b}}.

\bibitem[Cheng et~al.(2024)Cheng, Zhang, Yang, Nie, Li, Wu, and Shao]{cheng2024graphic}
Yutao Cheng, Zhao Zhang, Maoke Yang, Hui Nie, Chunyuan Li, Xinglong Wu, and Jie Shao.
\newblock Graphic design with large multimodal model.
\newblock \emph{arXiv preprint arXiv:2404.14368}, 2024.

\bibitem[Chng et~al.(2019)Chng, Liu, Sun, Ng, Luo, Ni, Fang, Zhang, Han, Ding, et~al.]{chng2019icdar2019}
Chee~Kheng Chng, Yuliang Liu, Yipeng Sun, Chun~Chet Ng, Canjie Luo, Zihan Ni, ChuanMing Fang, Shuaitao Zhang, Junyu Han, Errui Ding, et~al.
\newblock Icdar2019 robust reading challenge on arbitrary-shaped text-rrc-art.
\newblock In \emph{ICDAR}, 2019.

\bibitem[Daras and Dimakis(2022)]{daras2022discovering}
Giannis Daras and Alexandros~G Dimakis.
\newblock Discovering the hidden vocabulary of dalle-2.
\newblock \emph{arXiv preprint arXiv:2206.00169}, 2022.

\bibitem[Esser et~al.(2024)Esser, Kulal, Blattmann, Entezari, M{\"u}ller, Saini, Levi, Lorenz, Sauer, Boesel, et~al.]{esser2024scaling}
Patrick Esser, Sumith Kulal, Andreas Blattmann, Rahim Entezari, Jonas M{\"u}ller, Harry Saini, Yam Levi, Dominik Lorenz, Axel Sauer, Frederic Boesel, et~al.
\newblock Scaling rectified flow transformers for high-resolution image synthesis.
\newblock In \emph{ICML}, 2024.

\bibitem[flux(2024)]{flux}
flux.
\newblock Link: https://blackforestlabs.ai/\#get-flux, 2024.

\bibitem[Gao et~al.(2023)Gao, Lin, Zhou, Liu, Xie, Ge, and Jiang]{gao2023textpainter}
Yifan Gao, Jinpeng Lin, Min Zhou, Chuanbin Liu, Hongtao Xie, Tiezheng Ge, and Yuning Jiang.
\newblock Textpainter: Multimodal text image generation with visual-harmony and text-comprehension for poster design.
\newblock In \emph{ACMMM}, 2023.

\bibitem[Gupta et~al.(2021)Gupta, Lazarow, Achille, Davis, Mahadevan, and Shrivastava]{gupta2021layouttransformer}
Kamal Gupta, Justin Lazarow, Alessandro Achille, Larry~S Davis, Vijay Mahadevan, and Abhinav Shrivastava.
\newblock Layouttransformer: Layout generation and completion with self-attention.
\newblock In \emph{ICCV}, 2021.

\bibitem[He et~al.(2024)He, Liu, Chen, Tian, Liu, Chi, Liu, Yuan, Xing, Wang, et~al.]{he2024llms}
Yingqing He, Zhaoyang Liu, Jingye Chen, Zeyue Tian, Hongyu Liu, Xiaowei Chi, Runtao Liu, Ruibin Yuan, Yazhou Xing, Wenhai Wang, et~al.
\newblock Llms meet multimodal generation and editing: A survey.
\newblock \emph{arXiv preprint arXiv:2405.19334}, 2024.

\bibitem[Hertz et~al.(2023)Hertz, Mokady, Tenenbaum, Aberman, Pritch, and Cohen-Or]{hertz2022prompt}
Amir Hertz, Ron Mokady, Jay Tenenbaum, Kfir Aberman, Yael Pritch, and Daniel Cohen-Or.
\newblock Prompt-to-prompt image editing with cross attention control.
\newblock In \emph{ICLR}, 2023.

\bibitem[Horita et~al.(2024)Horita, Inoue, Kikuchi, Yamaguchi, and Aizawa]{horita2024retrieval}
Daichi Horita, Naoto Inoue, Kotaro Kikuchi, Kota Yamaguchi, and Kiyoharu Aizawa.
\newblock Retrieval-augmented layout transformer for content-aware layout generation.
\newblock In \emph{CVPR}, 2024.

\bibitem[Hsu et~al.(2023)Hsu, He, Peng, Kong, and Zhang]{hsu2023posterlayout}
Hsiao~Yuan Hsu, Xiangteng He, Yuxin Peng, Hao Kong, and Qing Zhang.
\newblock Posterlayout: A new benchmark and approach for content-aware visual-textual presentation layout.
\newblock In \emph{CVPR}, 2023.

\bibitem[Hu et~al.(2022)Hu, Shen, Wallis, Allen-Zhu, Li, Wang, Wang, and Chen]{hu2021lora}
Edward~J Hu, Yelong Shen, Phillip Wallis, Zeyuan Allen-Zhu, Yuanzhi Li, Shean Wang, Lu Wang, and Weizhu Chen.
\newblock Lora: Low-rank adaptation of large language models.
\newblock In \emph{ICLR}, 2022.

\bibitem[Inoue et~al.(2023)Inoue, Kikuchi, Simo-Serra, Otani, and Yamaguchi]{inoue2023towards}
Naoto Inoue, Kotaro Kikuchi, Edgar Simo-Serra, Mayu Otani, and Kota Yamaguchi.
\newblock Towards flexible multi-modal document models.
\newblock In \emph{CVPR}, 2023.

\bibitem[Inoue et~al.(2024)Inoue, Masui, Shimoda, and Yamaguchi]{inoue2024opencole}
Naoto Inoue, Kento Masui, Wataru Shimoda, and Kota Yamaguchi.
\newblock Opencole: Towards reproducible automatic graphic design generation.
\newblock \emph{arXiv preprint arXiv:2406.08232}, 2024.

\bibitem[Ji et~al.(2023)Ji, Zhang, Wang, Hou, Zhang, Price, and Chang]{ji2023improving}
Jiabao Ji, Guanhua Zhang, Zhaowen Wang, Bairu Hou, Zhifei Zhang, Brian Price, and Shiyu Chang.
\newblock Improving diffusion models for scene text editing with dual encoders.
\newblock \emph{Transactions on Machine Learning Research (TMLR)}, 2023.

\bibitem[Jia et~al.(2023)Jia, Li, Liu, Shen, Chen, Yuan, Zheng, Chen, Li, Xie, et~al.]{jia2023cole}
Peidong Jia, Chenxuan Li, Zeyu Liu, Yichao Shen, Xingru Chen, Yuhui Yuan, Yinglin Zheng, Dong Chen, Ji Li, Xiaodong Xie, et~al.
\newblock Cole: A hierarchical generation framework for graphic design.
\newblock \emph{arXiv preprint arXiv:2311.16974}, 2023.

\bibitem[Jia et~al.(2024)Jia, Yuan, Cheng, Wang, Li, Jia, and Zhang]{jia2024designedit}
Yueru Jia, Yuhui Yuan, Aosong Cheng, Chuke Wang, Ji Li, Huizhu Jia, and Shanghang Zhang.
\newblock Designedit: Multi-layered latent decomposition and fusion for unified \& accurate image editing.
\newblock \emph{arXiv preprint arXiv:2403.14487}, 2024.

\bibitem[Jiang et~al.(2022)Jiang, Sun, Zhu, Lou, and Zhang]{jiang2022coarse}
Zhaoyun Jiang, Shizhao Sun, Jihua Zhu, Jian-Guang Lou, and Dongmei Zhang.
\newblock Coarse-to-fine generative modeling for graphic layouts.
\newblock In \emph{AAAI}, 2022.

\bibitem[Jyothi et~al.(2019)Jyothi, Durand, He, Sigal, and Mori]{jyothi2019layoutvae}
Akash~Abdu Jyothi, Thibaut Durand, Jiawei He, Leonid Sigal, and Greg Mori.
\newblock Layoutvae: Stochastic scene layout generation from a label set.
\newblock In \emph{ICCV}, 2019.

\bibitem[Kikuchi et~al.(2024)Kikuchi, Inoue, Otani, Simo-Serra, and Yamaguchi]{kikuchi2024multimodal}
Kotaro Kikuchi, Naoto Inoue, Mayu Otani, Edgar Simo-Serra, and Kota Yamaguchi.
\newblock Multimodal markup document models for graphic design completion.
\newblock \emph{arXiv preprint arXiv:2409.19051}, 2024.

\bibitem[Kirillov et~al.(2023)Kirillov, Mintun, Ravi, Mao, Rolland, Gustafson, Xiao, Whitehead, Berg, Lo, et~al.]{kirillov2023segment}
Alexander Kirillov, Eric Mintun, Nikhila Ravi, Hanzi Mao, Chloe Rolland, Laura Gustafson, Tete Xiao, Spencer Whitehead, Alexander~C Berg, Wan-Yen Lo, et~al.
\newblock Segment anything.
\newblock In \emph{ICCV}, 2023.

\bibitem[Lan et~al.(2025)Lan, Bai, Duan, Li, Sun, and Chu]{lan2025flux}
Rui Lan, Yancheng Bai, Xu Duan, Mingxing Li, Lei Sun, and Xiangxiang Chu.
\newblock Flux-text: A simple and advanced diffusion transformer baseline for scene text editing.
\newblock \emph{arXiv preprint arXiv:2505.03329}, 2025.

\bibitem[Li et~al.(2021)Li, Zhang, and Wang]{li2021harmonious}
Chenhui Li, Peiying Zhang, and Changbo Wang.
\newblock Harmonious textual layout generation over natural images via deep aesthetics learning.
\newblock \emph{IEEE Transactions on Multimedia (TMM)}, 2021.

\bibitem[Li et~al.(2019)Li, Yang, Hertzmann, Zhang, and Xu]{li2019layoutgan}
Jianan Li, Jimei Yang, Aaron Hertzmann, Jianming Zhang, and Tingfa Xu.
\newblock Layoutgan: Generating graphic layouts with wireframe discriminators.
\newblock In \emph{ICLR}, 2019.

\bibitem[Li et~al.(2023)Li, Li, Feng, Zhu, Liu, Li, Zhang, Lv, Zhu, Shen, et~al.]{li2023planning}
Zhaochen Li, Fengheng Li, Wei Feng, Honghe Zhu, An Liu, Yaoyu Li, Zheng Zhang, Jingjing Lv, Xin Zhu, Junjie Shen, et~al.
\newblock Planning and rendering: Towards end-to-end product poster generation.
\newblock \emph{arXiv preprint arXiv:2312.08822}, 2023.

\bibitem[Liang et~al.(2024)Liang, Liu, Song, Jiang, Huang, Wang, and Li]{liang2024textcengen}
Tianyi Liang, Jiangqi Liu, Sicheng Song, Shiqi Jiang, Yifei Huang, Changbo Wang, and Chenhui Li.
\newblock Textcengen: Attention-guided text-centric background adaptation for text-to-image generation.
\newblock \emph{arXiv preprint arXiv:2404.11824}, 2024.

\bibitem[Lin et~al.(2023)Lin, Zhou, Ma, Gao, Fei, Chen, Yu, and Ge]{lin2023autoposter}
Jinpeng Lin, Min Zhou, Ye Ma, Yifan Gao, Chenxi Fei, Yangjian Chen, Zhang Yu, and Tiezheng Ge.
\newblock Autoposter: A highly automatic and content-aware design system for advertising poster generation.
\newblock In \emph{ACMMM}, 2023.

\bibitem[Liu et~al.(2023)Liu, Li, Wu, and Lee]{liu2024visual}
Haotian Liu, Chunyuan Li, Qingyang Wu, and Yong~Jae Lee.
\newblock Visual instruction tuning.
\newblock In \emph{NeurIPS}, 2023.

\bibitem[Liu et~al.(2024{\natexlab{a}})Liu, Liang, Liang, Luo, Li, Huang, and Yuan]{liu2024glyph}
Zeyu Liu, Weicong Liang, Zhanhao Liang, Chong Luo, Ji Li, Gao Huang, and Yuhui Yuan.
\newblock Glyph-byt5: A customized text encoder for accurate visual text rendering.
\newblock In \emph{ECCV}, 2024{\natexlab{a}}.

\bibitem[Liu et~al.(2024{\natexlab{b}})Liu, Liang, Zhao, Chen, Li, and Yuan]{liu2024glyph2}
Zeyu Liu, Weicong Liang, Yiming Zhao, Bohan Chen, Ji Li, and Yuhui Yuan.
\newblock Glyph-byt5-v2: A strong aesthetic baseline for accurate multilingual visual text rendering.
\newblock \emph{arXiv preprint arXiv:2406.10208}, 2024{\natexlab{b}}.

\bibitem[Ma et~al.(2023)Ma, Zhao, Chen, Wang, Niu, Lu, and Lin]{ma2023glyphdraw}
Jian Ma, Mingjun Zhao, Chen Chen, Ruichen Wang, Di Niu, Haonan Lu, and Xiaodong Lin.
\newblock Glyphdraw: Seamlessly rendering text with intricate spatial structures in text-to-image generation.
\newblock \emph{arXiv preprint arXiv:2303.17870}, 2023.

\bibitem[Ma et~al.(2024)Ma, Deng, Chen, Lu, and Yang]{ma2024glyphdraw2}
Jian Ma, Yonglin Deng, Chen Chen, Haonan Lu, and Zhenyu Yang.
\newblock Glyphdraw2: Automatic generation of complex glyph posters with diffusion models and large language models.
\newblock \emph{arXiv preprint arXiv:2407.02252}, 2024.

\bibitem[Podell et~al.(2023)Podell, English, Lacey, Blattmann, Dockhorn, M{\"u}ller, Penna, and Rombach]{podell2023sdxl}
Dustin Podell, Zion English, Kyle Lacey, Andreas Blattmann, Tim Dockhorn, Jonas M{\"u}ller, Joe Penna, and Robin Rombach.
\newblock Sdxl: Improving latent diffusion models for high-resolution image synthesis.
\newblock In \emph{ICLR}, 2023.

\bibitem[Pu et~al.(2025)Pu, Zhao, Tang, Yin, Ye, Yuan, Chen, Bao, Zhang, Wang, et~al.]{pu2025art}
Yifan Pu, Yiming Zhao, Zhicong Tang, Ruihong Yin, Haoxing Ye, Yuhui Yuan, Dong Chen, Jianmin Bao, Sirui Zhang, Yanbin Wang, et~al.
\newblock Art: Anonymous region transformer for variable multi-layer transparent image generation.
\newblock In \emph{CVPR}, 2025.

\bibitem[Rombach et~al.(2022)Rombach, Blattmann, Lorenz, Esser, and Ommer]{rombach2022high}
Robin Rombach, Andreas Blattmann, Dominik Lorenz, Patrick Esser, and Bj{\"o}rn Ommer.
\newblock High-resolution image synthesis with latent diffusion models.
\newblock In \emph{CVPR}, 2022.

\bibitem[Seol et~al.(2024)Seol, Kim, and Yoo]{seol2024posterllama}
Jaejung Seol, Seojun Kim, and Jaejun Yoo.
\newblock Posterllama: Bridging design ability of langauge model to contents-aware layout generation.
\newblock \emph{arXiv preprint arXiv:2404.00995}, 2024.

\bibitem[Shabani et~al.(2024)Shabani, Wang, Liu, Zhao, Yang, and Furukawa]{shabani2024visual}
Mohammad~Amin Shabani, Zhaowen Wang, Difan Liu, Nanxuan Zhao, Jimei Yang, and Yasutaka Furukawa.
\newblock Visual layout composer: Image-vector dual diffusion model for design layout generation.
\newblock In \emph{CVPR}, 2024.

\bibitem[Shimoda et~al.(2021)Shimoda, Haraguchi, Uchida, and Yamaguchi]{shimoda2021rendering}
Wataru Shimoda, Daichi Haraguchi, Seiichi Uchida, and Kota Yamaguchi.
\newblock De-rendering stylized texts.
\newblock In \emph{ICCV}, 2021.

\bibitem[Tudosiu et~al.(2024)Tudosiu, Yang, Zhang, Chen, McDonagh, Lampouras, Iacobacci, and Parisot]{tudosiu2024mulan}
Petru-Daniel Tudosiu, Yongxin Yang, Shifeng Zhang, Fei Chen, Steven McDonagh, Gerasimos Lampouras, Ignacio Iacobacci, and Sarah Parisot.
\newblock Mulan: A multi layer annotated dataset for controllable text-to-image generation.
\newblock In \emph{CVPR}, 2024.

\bibitem[Tuo et~al.(2024)Tuo, Xiang, He, Geng, and Xie]{tuo2023anytext}
Yuxiang Tuo, Wangmeng Xiang, Jun-Yan He, Yifeng Geng, and Xuansong Xie.
\newblock Anytext: Multilingual visual text generation and editing.
\newblock In \emph{ICLR}, 2024.

\bibitem[Xu et~al.(2023)Xu, Wang, Zhang, Wang, and Shi]{xu2023versatile}
Xingqian Xu, Zhangyang Wang, Gong Zhang, Kai Wang, and Humphrey Shi.
\newblock Versatile diffusion: Text, images and variations all in one diffusion model.
\newblock In \emph{ICCV}, 2023.

\bibitem[Yamaguchi(2021)]{yamaguchi2021canvasvae}
Kota Yamaguchi.
\newblock Canvasvae: Learning to generate vector graphic documents.
\newblock In \emph{ICCV}, 2021.

\bibitem[Yu et~al.(2022)Yu, Chen, Chen, Meng, Wu, Josel, Niebles, Xiong, and Xu]{yu2022layoutdetr}
Ning Yu, Chia-Chih Chen, Zeyuan Chen, Rui Meng, Gang Wu, Paul Josel, Juan~Carlos Niebles, Caiming Xiong, and Ran Xu.
\newblock Layoutdetr: detection transformer is a good multimodal layout designer.
\newblock \emph{arXiv preprint arXiv:2212.09877}, 2022.

\bibitem[Zeng et~al.(2024)Zeng, Shu, Li, Yang, and Zhou]{zeng2024textctrl}
Weichao Zeng, Yan Shu, Zhenhang Li, Dongbao Yang, and Yu Zhou.
\newblock Textctrl: Diffusion-based scene text editing with prior guidance control.
\newblock In \emph{NeurIPS}, 2024.

\bibitem[Zhang et~al.(2023{\natexlab{a}})Zhang, Guo, Sun, Lou, and Zhang]{zhang2023layoutdiffusion}
Junyi Zhang, Jiaqi Guo, Shizhao Sun, Jian-Guang Lou, and Dongmei Zhang.
\newblock Layoutdiffusion: Improving graphic layout generation by discrete diffusion probabilistic models.
\newblock In \emph{ICCV}, 2023{\natexlab{a}}.

\bibitem[Zhang and Agrawala(2024)]{zhang2024transparent}
Lvmin Zhang and Maneesh Agrawala.
\newblock Transparent image layer diffusion using latent transparency.
\newblock \emph{ACM TRANSACTIONS ON GRAPHICS (TOG)}, 2024.

\bibitem[Zhang et~al.(2023{\natexlab{b}})Zhang, Rao, and Agrawala]{zhang2023adding}
Lvmin Zhang, Anyi Rao, and Maneesh Agrawala.
\newblock Adding conditional control to text-to-image diffusion models.
\newblock In \emph{ICCV}, 2023{\natexlab{b}}.

\bibitem[Zhang et~al.(2024)Zhang, Chen, Wang, Lu, and Qiao]{zhang2024brush}
Lingjun Zhang, Xinyuan Chen, Yaohui Wang, Yue Lu, and Yu Qiao.
\newblock Brush your text: Synthesize any scene text on images via diffusion model.
\newblock In \emph{AAAI}, 2024.

\bibitem[Zhang et~al.(2023{\natexlab{c}})Zhang, Zhao, Lu, and Chien]{zhang2023text2layer}
Xinyang Zhang, Wentian Zhao, Xin Lu, and Jeff Chien.
\newblock Text2layer: Layered image generation using latent diffusion model.
\newblock \emph{arXiv preprint arXiv:2307.09781}, 2023{\natexlab{c}}.

\bibitem[Zhangli et~al.(2024)Zhangli, Jiang, Liu, Yu, Dai, Ramchandani, Pang, Metaxas, and Krishnan]{zhangli2024layout}
Qilong Zhangli, Jindong Jiang, Di Liu, Licheng Yu, Xiaoliang Dai, Ankit Ramchandani, Guan Pang, Dimitris~N Metaxas, and Praveen Krishnan.
\newblock Layout-agnostic scene text image synthesis with diffusion models.
\newblock In \emph{CVPR}, 2024.

\bibitem[Zhao et~al.(2018)Zhao, Cao, and Lau]{zhao2018modeling}
Nanxuan Zhao, Ying Cao, and Rynson~WH Lau.
\newblock Modeling fonts in context: Font prediction on web designs.
\newblock In \emph{Computer Graphics Forum}, 2018.

\bibitem[Zhao and Lian(2024)]{zhao2023udifftext}
Yiming Zhao and Zhouhui Lian.
\newblock Udifftext: A unified framework for high-quality text synthesis in arbitrary images via character-aware diffusion models.
\newblock In \emph{ECCV}, 2024.

\bibitem[Zhao et~al.(2024)Zhao, Tang, Wu, Lin, Wei, Liu, Tan, Zhang, Huang, and Xie]{zhao2024harmonizing}
Zhen Zhao, Jingqun Tang, Binghong Wu, Chunhui Lin, Shu Wei, Hao Liu, Xin Tan, Zhizhong Zhang, Can Huang, and Yuan Xie.
\newblock Harmonizing visual text comprehension and generation.
\newblock In \emph{NeurIPS}, 2024.

\bibitem[Zheng et~al.(2019)Zheng, Qiao, Cao, and Lau]{zheng2019content}
Xinru Zheng, Xiaotian Qiao, Ying Cao, and Rynson~WH Lau.
\newblock Content-aware generative modeling of graphic design layouts.
\newblock \emph{ACM Transactions on Graphics (TOG)}, 2019.

\bibitem[Zhou et~al.(2022)Zhou, Xu, Ma, Ge, Jiang, and Xu]{zhou2022composition}
Min Zhou, Chenchen Xu, Ye Ma, Tiezheng Ge, Yuning Jiang, and Weiwei Xu.
\newblock Composition-aware graphic layout gan for visual-textual presentation designs.
\newblock \emph{arXiv preprint arXiv:2205.00303}, 2022.

\bibitem[Zhu et~al.(2024)Zhu, Healey, Zhang, Wang, and Sun]{zhu2024automatic}
Wanrong Zhu, Jennifer Healey, Ruiyi Zhang, William~Yang Wang, and Tong Sun.
\newblock Automatic layout planning for visually-rich documents with instruction-following models.
\newblock \emph{arXiv preprint arXiv:2404.15271}, 2024.

\end{thebibliography}
}

\clearpage

\onecolumn

\appendix

\noindent{\LARGE{\textbf{Appendix}}}


\section{Gallery for more text-to-template results}
\begin{figure}[ht!]
    \centering
    \includegraphics[width=0.4\textwidth]{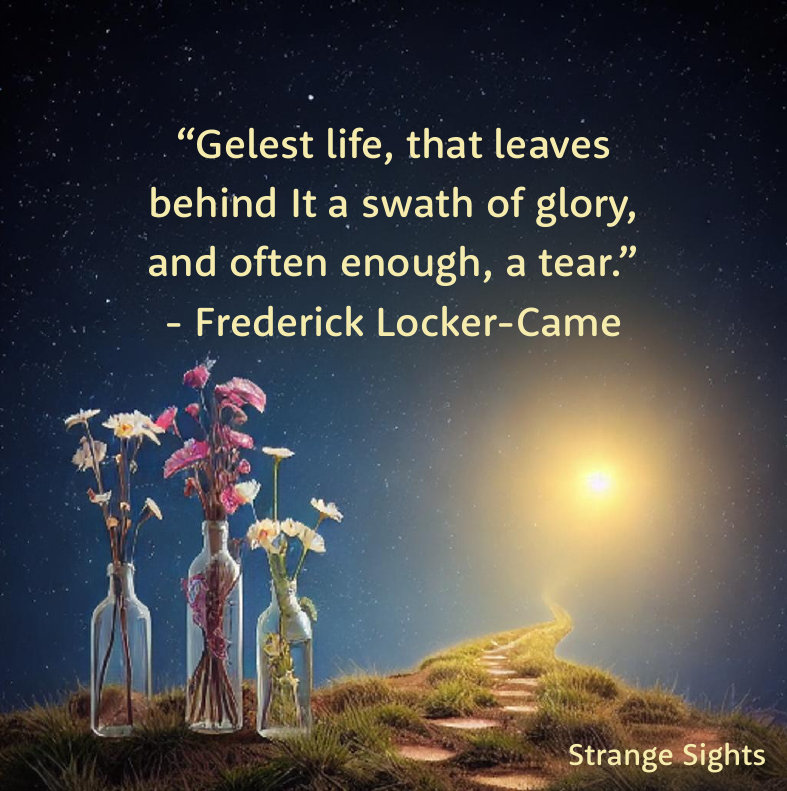}
    \hfill 
    \includegraphics[width=0.4\textwidth]{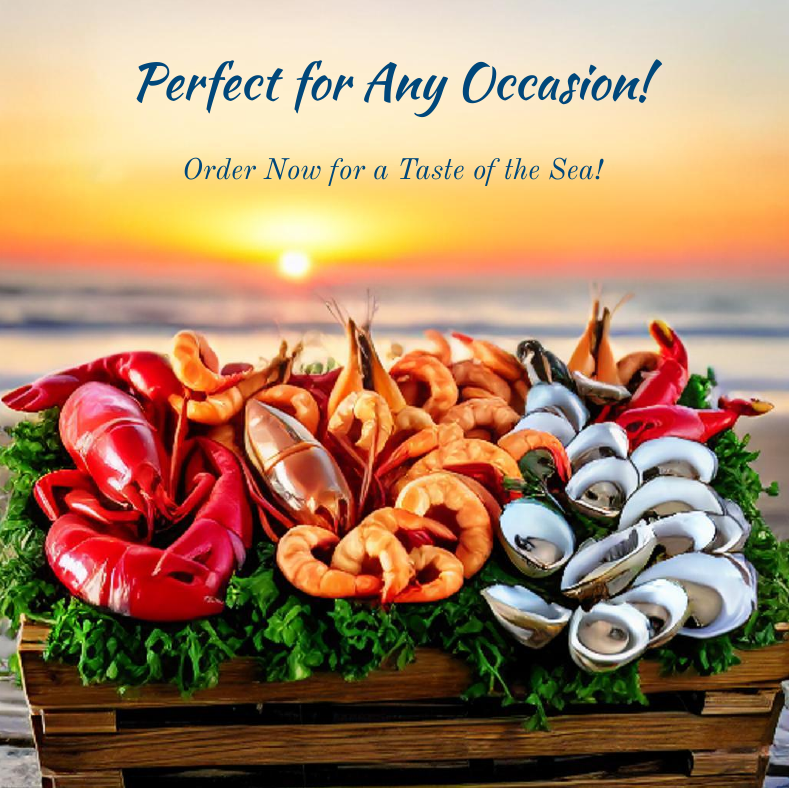}
    

    \includegraphics[width=0.4\textwidth]{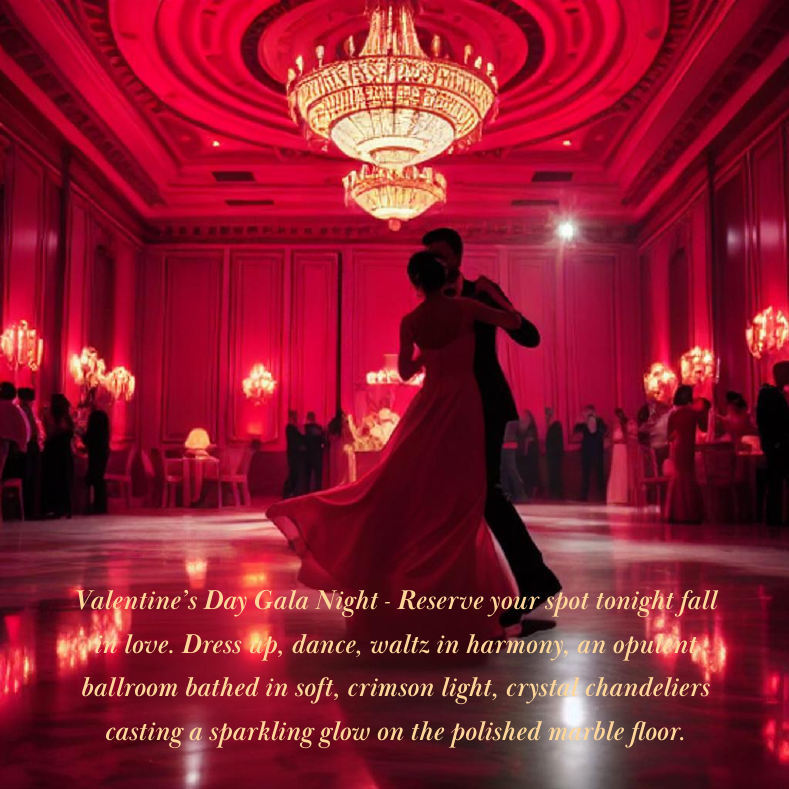}
    \hfill
    \includegraphics[width=0.4\textwidth]{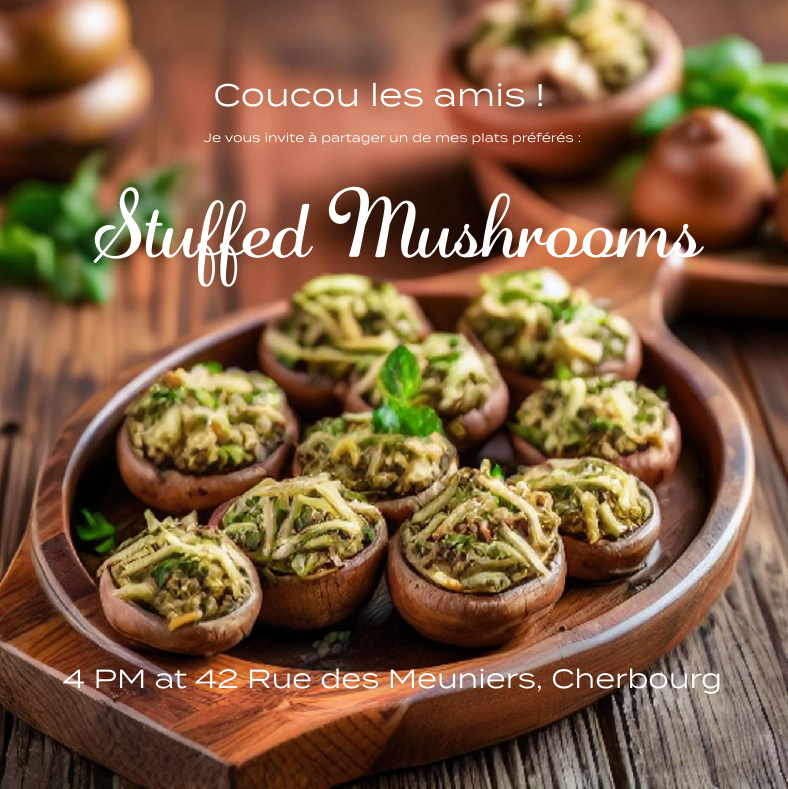}

    \includegraphics[width=0.4\textwidth]{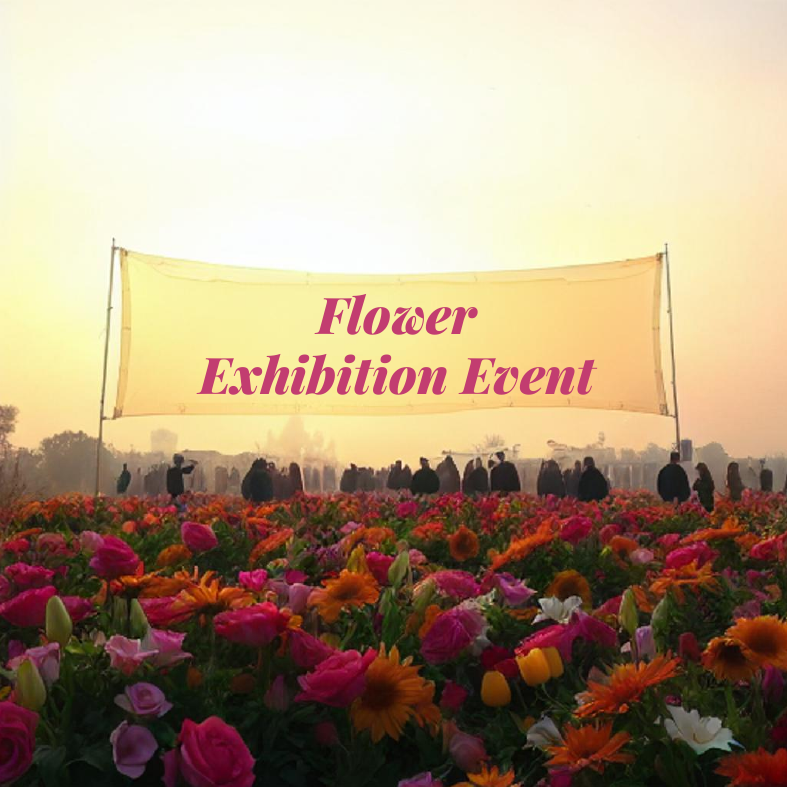}
    \hfill
    \includegraphics[width=0.4\textwidth]{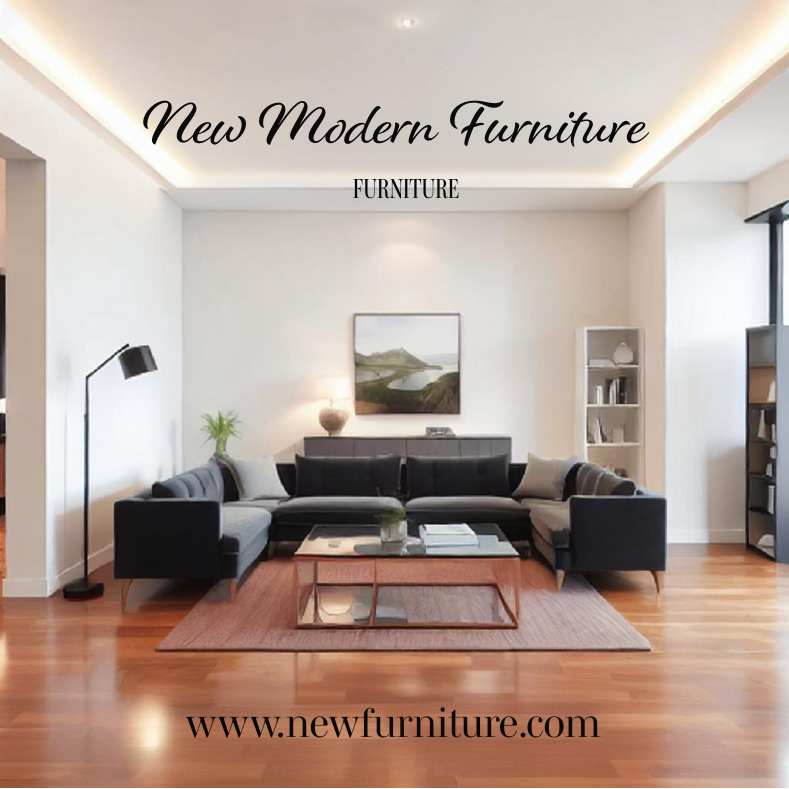}

    \caption{Visualizations of more text-to-template results. [1/2]}
    \label{fig:gallery1}
\end{figure}

\begin{figure}[ht!]
    \centering
    \includegraphics[width=0.4\textwidth]{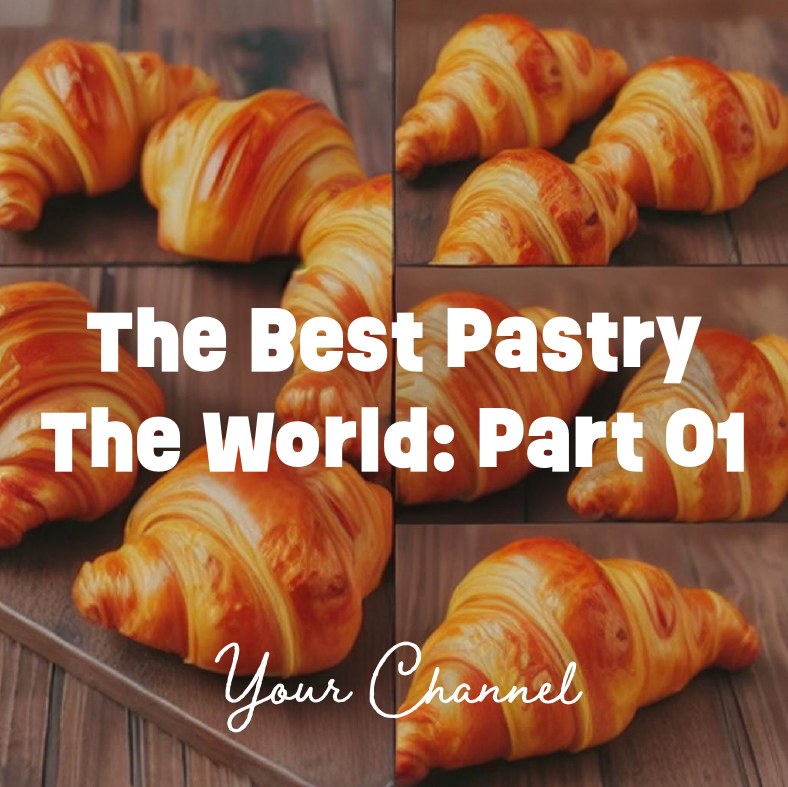}
    \hfill 
    \includegraphics[width=0.4\textwidth]{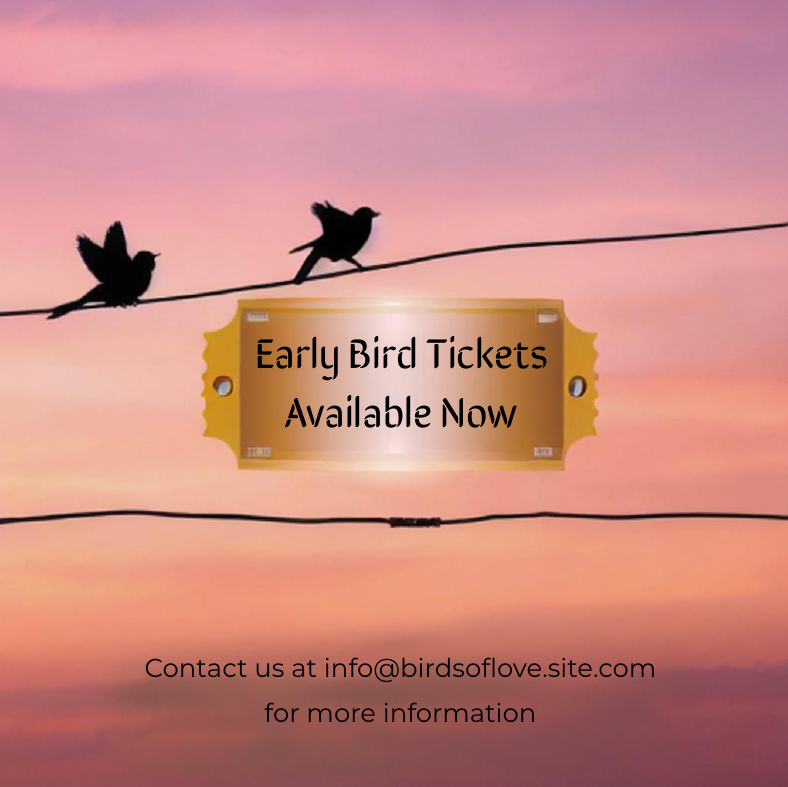}
    

    \includegraphics[width=0.4\textwidth]{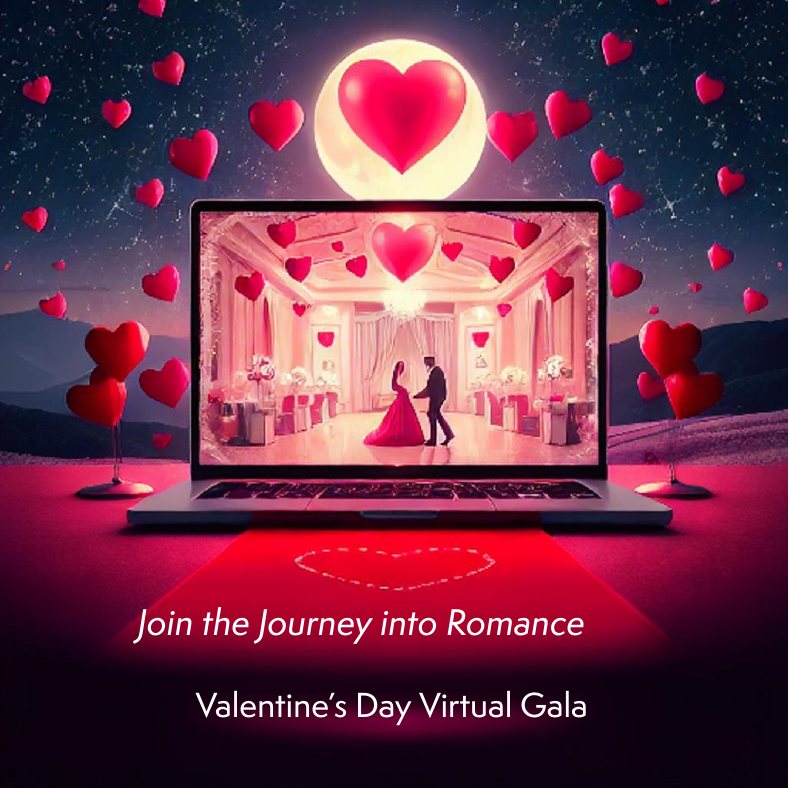}
    \hfill
    \includegraphics[width=0.4\textwidth]{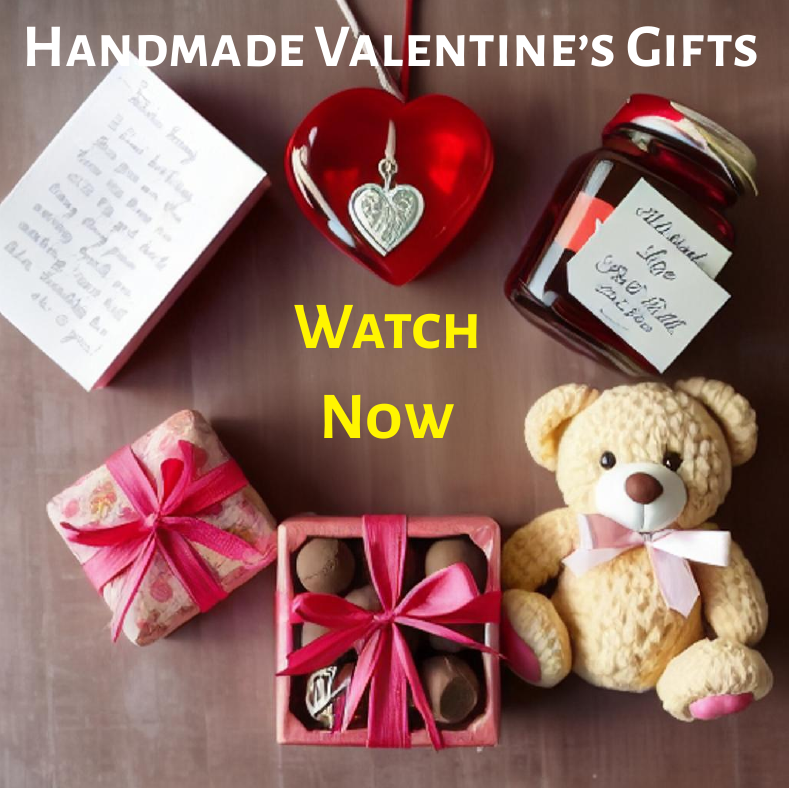}

    \includegraphics[width=0.4\textwidth]{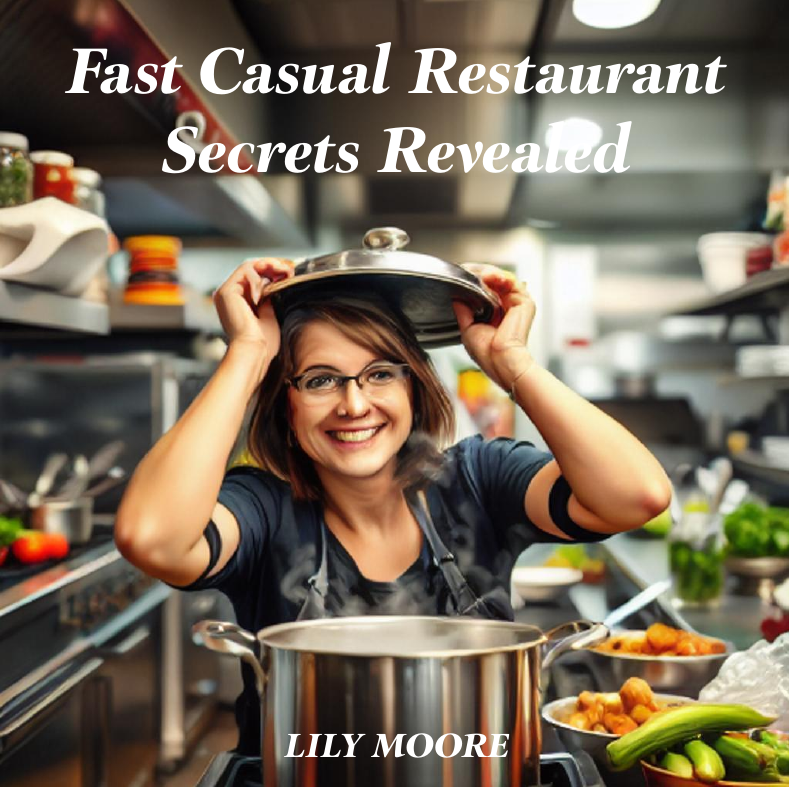}
    \hfill
    \includegraphics[width=0.4\textwidth]{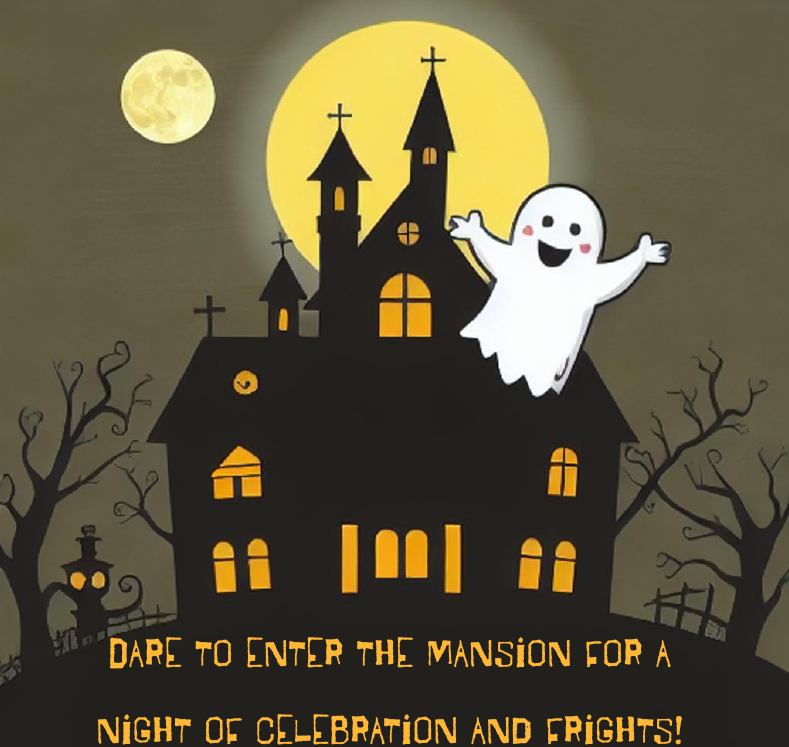}

    \caption{Visualizations of more text-to-template results. [2/2]}
    \label{fig:gallery2}
\end{figure}

We demonstrate more text-to-template results in Figure \ref{fig:gallery1} and Figure \ref{fig:gallery2}. Due to the text being layered and vectorized, operations such as selection and copying are feasible.

\section{Prompts for the stage1 reference creation}

Below we use in-context learning to make VLM expand the given prompt:

\begin{mytextbox}
Your task is to expand the original prompt into a detailed one. I will give you some examples.

[Input 1]
Create an advertisement for a fish market with a special offer of a 20\% discount on seafood.

[Output 1]
A bustling fish market under a vibrant morning sky. Local vendors display an array of fresh, glistening seafood, from ruby-red lobsters to iridescent, silver fish. A large, colorful banner hangs overhead, proudly announcing a special offer with 20\% discount on all seafood. The air is thick with excitement and the irresistible aroma of the ocean.

[Input 2]
Create a business card for a flower shop with a focus on blue tulips.

[Output 2]
An elegant business card lying on a white marble surface. The card is adorned with a captivating watercolor illustration of rich, azure blue tulips, their petals opening up to reveal layers of deep and light shades of blue. The shop's name is written in a sophisticated cursive font at the center, while contact details are subtly placed at the bottom right corner.

Now based on the given prompt "Design a cutting-edge logo for a real estate agency named Golden Home.", please expand it into a detailed one.
\end{mytextbox}

When users wish to provide additional constraints, they can input sketches to the VLM. We employ the following task prompt template to query the VLM:

\begin{mytextbox}
You will be provided with a sketch that you need to analyze and describe meticulously, paying close attention to each detail depicted. Identify and describe where each object is located within the sketch. Note that the "xxx" symbols on the image are placeholders for text, which you should replace with appropriate content. Your description should capture the layout and the thematic elements of the design. As a reference information, this image is about "eating more apples is good for your health". 
\end{mytextbox}

\begin{figure}[h]
\centering
\includegraphics[width=1.0\textwidth]{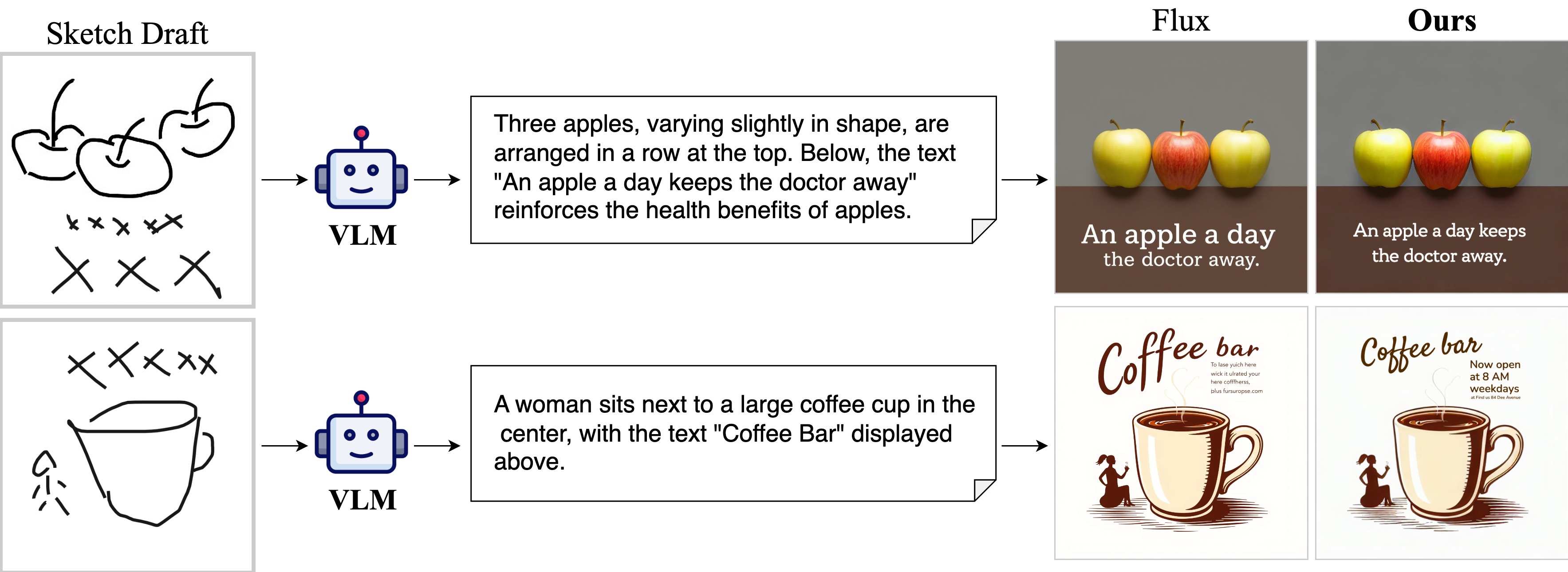}
\caption{The VLM can convert the sketch draft into detailed prompt to generate references.}
\label{fig:sketch}
\end{figure}

This template explicitly describes the task and indicates that specific text should replace the placeholder `xxx'. It also includes a brief user intention, providing the VLM with an overview of the image. We demonstrate some cases in Figure \ref{fig:sketch}.

\section{Combined prompts for the stage2 design planning}

We showcase some examples in the below. For training GenAI design, we use the following template:

\begin{mytextbox}
Parse and refine the attributions of text. Parse the objects, and backgrounds in the graphic design image. The caption of the image is The "Red White Bold Type" beverage label is a striking visual feast, designed to capture the essence of boldness and purity. With a vivid red and pristine white color scheme, the label features bold, assertive typography that commands attention. This design not only reflects the vibrant and robust flavors of the beverage but also appeals to consumers with its clean, contemporary aesthetic, making it a standout choice on any shelf. Support OCR results are: [[(22, 64, 228, 132)], [(21, 126, 311, 211)], [(82, 208, 119, 215)]].
\end{mytextbox}

\begin{mytextbox}
Parse and refine the attributions of text. Parse the objects, and backgrounds in the graphic design image. The caption of the image is The Facebook page cover for a modern record store should be a vibrant and engaging visual that encapsulates the essence of music and contemporary design. It might feature a collage of iconic album covers, interspersed with sleek, modern graphic elements that convey the store's cutting-edge aesthetic. Support OCR results are: [[(214, 89, 299, 120)], [(41, 86, 110, 138)], [(18, 121, 59, 176)], [(195, 121, 317, 147)], [(209, 175, 310, 197)], [(224, 197, 290, 219)], [(84, 219, 106, 237)], [(215, 232, 300, 246)]].
\end{mytextbox}

For training the original design, we use the following template. Please note that here we do not incorporate the description since the text within the design already contains massive information. Meanwhile, we integrate the OCR recognition result in the OCR string.

\begin{mytextbox}
Parse the attributions of text, objects, and backgrounds in the graphic design image. Support OCR results are: [['THE COOD', (85, 15, 228, 51)], ['CREATIVE', (88, 51, 232, 85)], ['STUDIO', (84, 83, 196, 120)], ['2701Willow', (85, 218, 158, 236)], ['Charles,', (83, 235, 135, 253)], ['aneLake', (122, 228, 177, 243)], ['(555)555-0100', (85, 265, 174, 282)], ['@thegoodstudio', (86, 297, 180, 312)], ['www.thegoodstudio.site.con', (85, 310, 237, 324)]]
\end{mytextbox}

\begin{mytextbox}
Parse the attributions of text, objects, and backgrounds in the graphic design image. Support OCR results are: [['CLEARANCE', (19, 213, 318, 255)], ['SALE', (14, 256, 136, 297)], ['2701WillowOaks', (203, 272, 300, 287)], ['Lane Lake Charles,LA', (203, 284, 321, 298)]]
\end{mytextbox}

For training the backgrounds with text, we use the following template. Note that the OCR string is omitted since there is no text within the design.

\begin{mytextbox}
Add text on the background. And parse the overall graphic design. The caption of the image is floral green and pink wellness institute business card.
\end{mytextbox}

\begin{mytextbox}
Add text on the background. And parse the overall graphic design. The caption of the image is The logo for Green Saw Carpenters captures the essence of the brands commitment to sustainable building practices and skilled craftsmanship. It features a stylized green saw blade, intricately designed to resemble both a leaf and a carpentry tool, symbolizing the fusion of nature and construction.
\end{mytextbox}

\section{More details about the questionnaire for result selection}

\begin{figure}[h]
\centering
\includegraphics[width=1.0\textwidth]{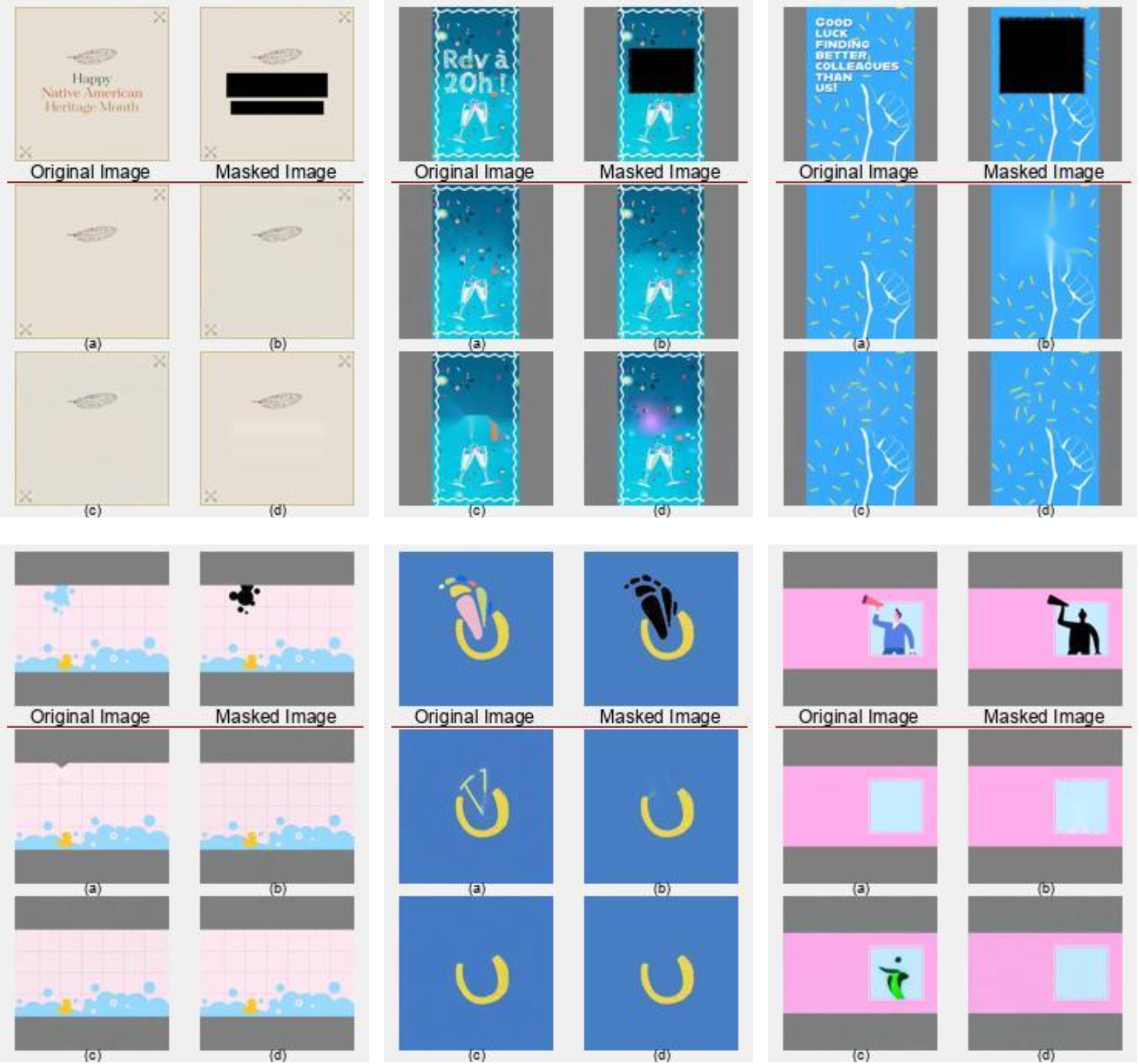}
\caption{More examples about the questionnaire. The samples in the first row is for text removal, and the samples in the second row is for object removal.}
\label{fig:questionnaire}
\end{figure}

We show the task prompt in the following, and display some cases in Figure \ref{fig:questionnaire}.

\begin{mytextbox}
The provided image appears to show four different results of a graphic design removal task. The first row displays the original image on the left and the masked image on the right. The second and third rows exhibit the corresponding outcomes of the graphic design removal. To evaluate the effectiveness of the results, the key criteria are: 1) the overall harmony and coherence of the image, 2) the purity and cleanness of the background, and 3) the absence of any additional, extraneous elements. Based on these criteria, please select the option (a, b, c, or d) that represents the best result.
\end{mytextbox}

\section{More details about Design39K}

Here we provide additional statistics about the in-house Design39K dataset. The average output sequence length is 728.32. To ensure that the majority of cases do not exceed this limit, we have set the maximum output length for the VLM to 1,536. Each design, on average, contains about 1.02 objects and 3.11 text regions. Notably, the dataset utilizes the title of each design as the description.

\section{More details about each vision expert}

\noindent \textbf{Flux.} We use the open-source Flux.1 schnell to create the reference. We observe that, even without explicit character guidance like other methods \cite{chen2023textdiffuser,tuo2023anytext}, Flux still synthesizes high-quality design references. Flux demonstrates robust performance in generating reference images compared with previous methods such as SD 1.5 \cite{rombach2022high}. In some cases, Flux may struggle to generate design images according to the prompts, instead producing backgrounds without text. In these situations, users may need to attempt multiple times to obtain the desired reference image. The output size of the model is 1024$\times$1024. The number of sampling steps is set to 4, and the maximum length of the input prompt is set to 256, both as default.

\noindent \textbf{PaddleOCR.} PaddleOCR is an open-source optical character recognition toolkit that provides practical and efficient solutions for text detection and recognition across various images. It has been observed that PaddleOCR also exhibits strong detection performance on GenAI images.

\noindent \textbf{SAM.} Segment Anything Model \cite{kirillov2023segment} is an advanced segmentation model designed to perform highly accurate and versatile image segmentation across an extensive array of objects and scenes, enabling detailed and automated analysis of visual data. Here we employ the detection box to obtain the segmentation mask. Specifically, we employ the ``sam-vit-base'' architecture to get the segmentation mask. 

\noindent \textbf{Removal model.} We achieve object removal results using the ControlNet inpainting model based on SD 1.5 \cite{rombach2022high}, employing the prompt ``nothing in the image'' to erase specific content. The input size and the output size are 512$\times$512. While we notice that a few samples exhibit color shifting, we consider this acceptable as the results still appear harmonious.

\section{Details about the Streamlit frontend}
We develop a Streamlit-based front-end system to facilitate the presentation and manipulation of layered graphic designs. This system enables the separation and individual rendering of various design elements, including text, images, and objects, thereby allowing for flexible control and real-time previewing of design components. Specifically, we leverage HTML and CSS to render text elements and employ the st.elements.image component to display images and objects.

\section{Capability to produce non-square layered design}
\begin{figure}[ht!]
    \centering
    
    \includegraphics[width=0.4\textwidth]{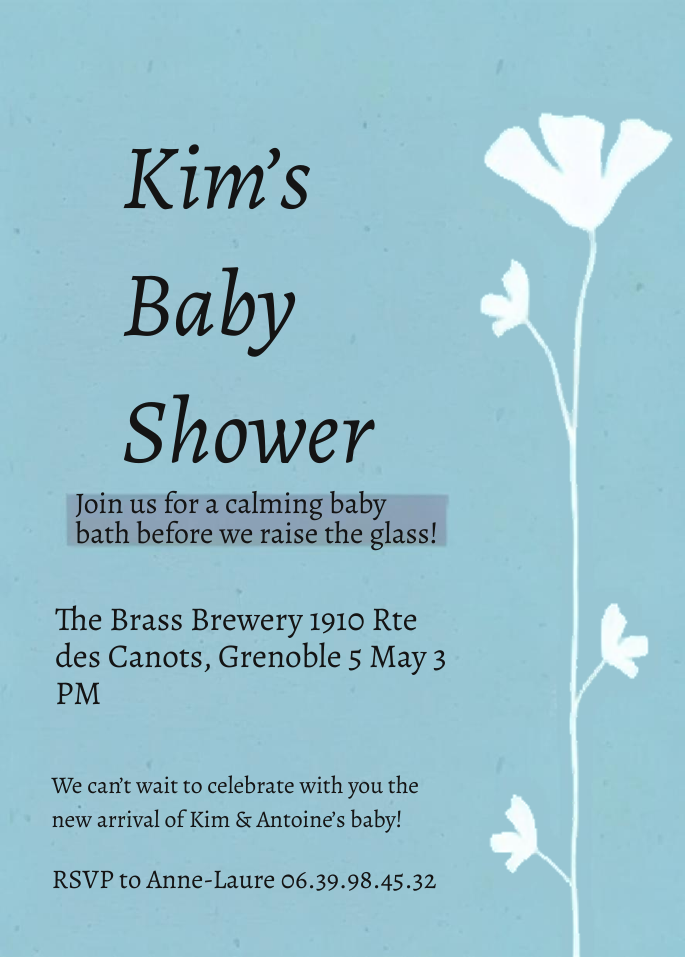}
    \hfill
    \includegraphics[width=0.4\textwidth]{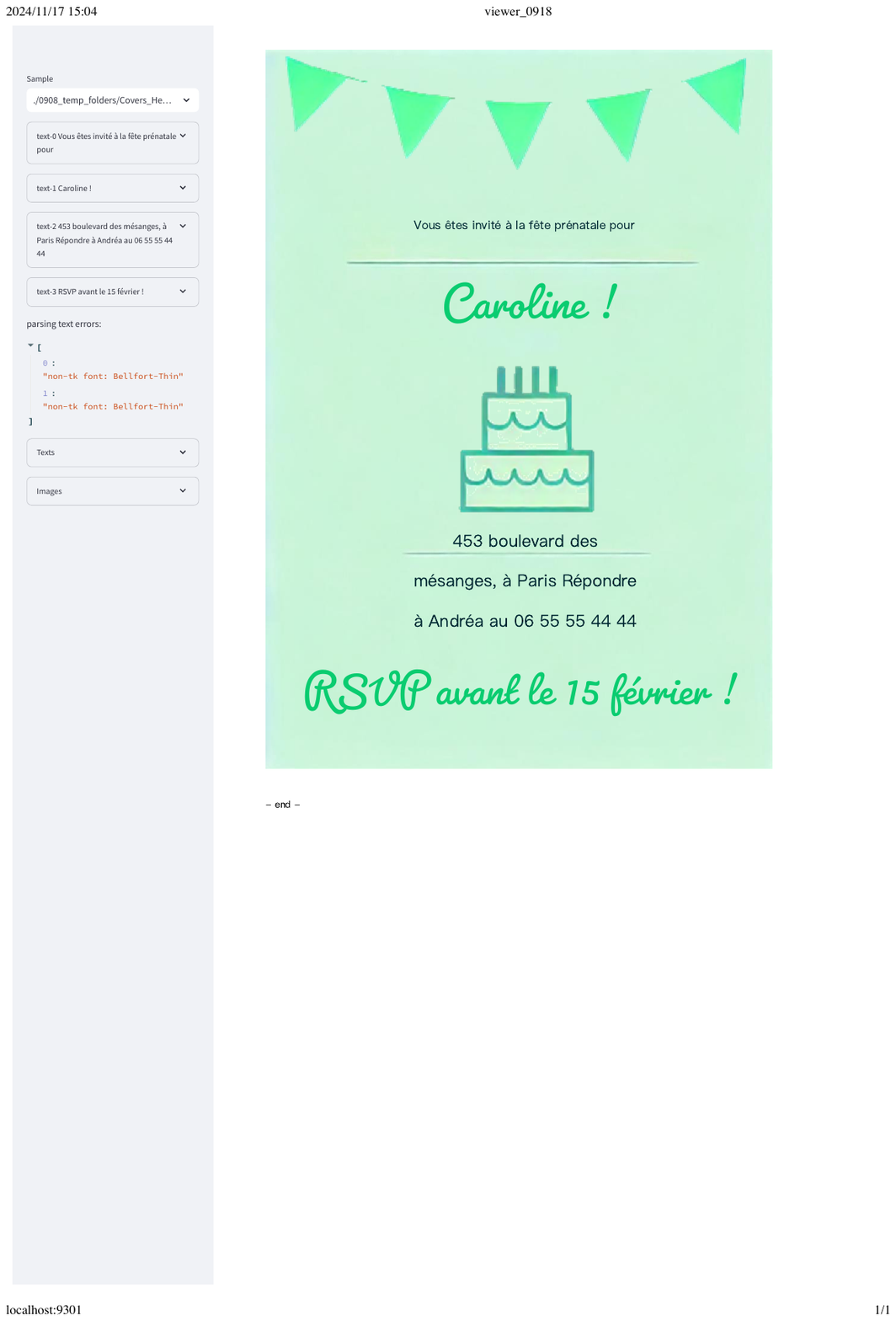}

    \caption{Visualizations of non-square designs.}
    \label{fig:non-square}
\end{figure}

Please note that the proposed Accordion method can produce non-layered designs, which sets it apart from existing methods like COLE \cite{jia2023cole} and Open-COLE \cite{inoue2024opencole} that are confined to layered designs. For instance, when a non-square design is used as a reference in the second stage, it is initially padded with gray borders to facilitate the design planning process. Subsequently, the SAM and removal models are applied to the padded image. Finally, the borders are deleted to yield the non-square designs. The results are shown in Figure \ref{fig:non-square}.

\section{Task prompt for evaluating adding text to background}

\begin{figure}[h]
\centering
\includegraphics[width=0.7\textwidth]{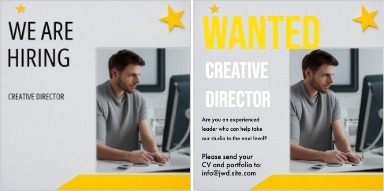}
\caption{Samples generated by COLE and ours are concatenated for comparison.}
\label{fig:questionnaire2}
\end{figure}

Here we use the following prompt for GPT-4V. The horizontal concatenation is demonstrated in Figure \ref{fig:questionnaire2}.

\begin{mytextbox}
Here you will see two designs with the same background but different text content and placement. Please consider the factors such as layout, content relevance, typography and color scheme. Select which one is better, you can answer the left one is better / the right one is better, then detail the reasons.
\end{mytextbox}

\section{Details of user studies by designers}
\begin{figure}[h]
\centering
\includegraphics[width=1\textwidth]{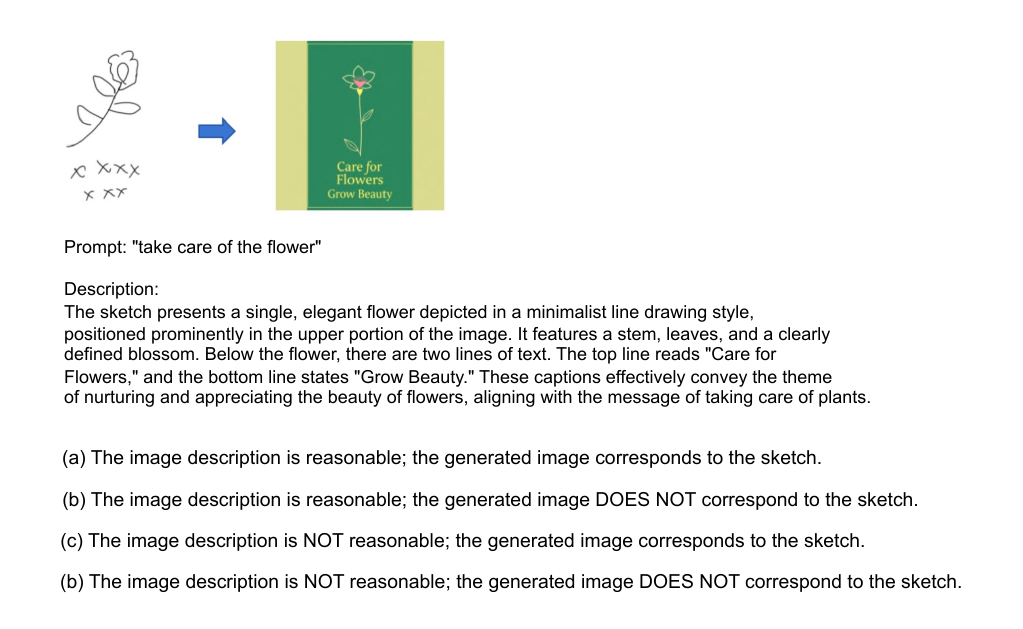}
\caption{User studies by designers to assess to task of sketch to design images.}
\label{fig:userstudy}
\end{figure}

\begin{figure}[h]
\centering
\includegraphics[width=1\textwidth]{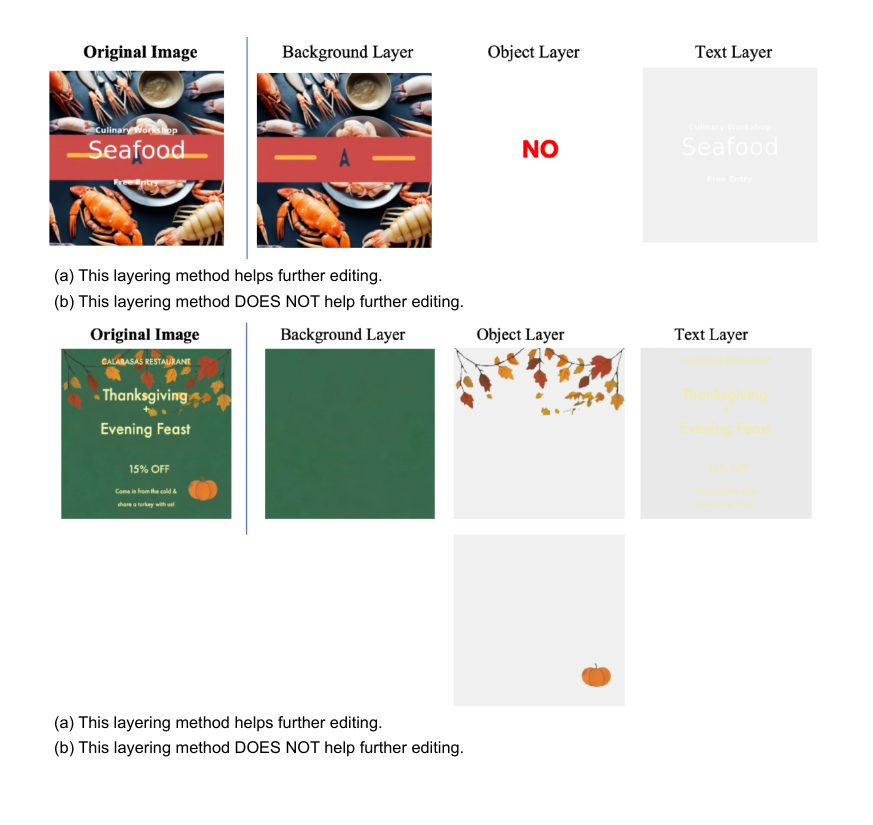}
\caption{User studies by designers to assess to task of text-to-templates.}
\label{fig:userstudy2}
\end{figure}

In our user studies, we present several cases to evaluate the effectiveness of converting sketches to design images (Figure \ref{fig:userstudy}) and text to templates \ref{fig:userstudy2}. We specifically invite individuals with design backgrounds to participate in these studies, ensuring that the assessments are informed by professional insights. This approach enables us to gather valuable feedback on the usability and accuracy of our models from experts in the field, enhancing the validity of our study results.

\section{Details about the experiment on the benefits of joint training}

In Table \ref{tab:details}, we present the evaluation results using various types of images as inputs. In Table \ref{tab:details}, we present the evaluation results using various types of images as inputs. Note that we use normalized edit distance (NED) for evaluating OCR accuracy, which is particularly effective when our OCR is applied to paragraph-level long text following \cite{chng2019icdar2019}. For color accuracy, we consider the predictions correct only when the values for red (R), green (G), blue (B), and alpha (A) channels are exactly accurate. When the input images are designs with nonsensical text or backgrounds without text, it becomes challenging to assess text-related metrics. Therefore, we rely solely on object-related metrics for evaluation. The results indicate that the scores for both categories are comparable. Given our goal to develop a compact model, we ultimately opt for joint training.

We also evaluate the effect of integrating an OCR prompt into our system. For the task of text recognition aimed at parsing the original design, we note a 7.23\% improvement in paragraph-level OCR Normalized Edit Distance (NED), increasing from 61.28\% to 68.51\%. Additionally, the average detection F1 score for the text detection task in both the original and GenAI designs has improved by 5.46\%, rising from 73.12\% to 78.59\%.

\begin{table}[h]
\centering
\caption{Ablation studies about the experiment on the benefits of joint training.}
\label{tab:details}
\begin{tabular}{lcc}
\toprule
Metrics & Separate Training & Joint Training \\
\midrule
\textbf{\textit{Original Design}} & \\
 Text Detection F1 & 75.42 & \textbf{78.59} \\
 Text Recognition NED & \textbf{72.87} & 68.51 \\
 Object Detection F1 & 82.17 & \textbf{84.64} \\
 Color Accuracy & 26.66 & \textbf{28.09} \\
 Font Accuracy & \textbf{24.51} & 21.62 \\
 Line Number Accuracy & \textbf{86.96} & 86.28 \\
 Alignment Accuracy & 87.28 & \textbf{88.60} \\
 Angle Accuracy & 90.16 & \textbf{91.52} \\
\midrule
\textbf{\textit{Designs with Nonsensical Text}} & \\
 Object Detection F1 & 79.27 & \textbf{83.06} \\
\midrule
\textbf{\textit{Backgrounds without Text}} & \\
 Object Detection F1 & 83.52 & \textbf{86.94} \\
 \midrule
\textbf{\textit{Questionnaire Result Selection}} & \\
 Selection Accuracy & \textbf{83.54} & \textbf{83.54} 
\\
 \midrule
\underline{\textbf{\textit{Average Score}}} & 72.03 & \textbf{72.85} 
\\
\bottomrule
\end{tabular}
\end{table}

\end{document}